\documentclass[a4paper,fleqn]{cas-sc}

\usepackage[authoryear,longnamesfirst]{natbib}
\usepackage{hyperref}
\usepackage{float}
\usepackage{xcolor}
\usepackage{booktabs}
\usepackage{multirow}

\usepackage{verbatim} 
\usepackage{apalike}
\usepackage{amsmath, amssymb}
\restylefloat{figure}
\floatstyle{plaintop} 
\restylefloat{table}
\def\tsc#1{\csdef{#1}{\textsc{\lowercase{#1}}\xspace}}
\tsc{WGM}
\tsc{QE}
\tsc{EP}
\tsc{PMS}
\tsc{BEC}
\tsc{DE}

\begin{document}
\let\WriteBookmarks\relax
\def\floatpagepagefraction{1}
\def\textpagefraction{.001}
\shorttitle{}
\shortauthors{B. Ispizua et~al.}

\title [mode = title]{Real-Time State-of-Health Estimation and Online Degradation Prognosis from Partial Battery Discharge Using Physics-Informed Neural Networks}                      
\tnotemark[1]

\tnotetext[1]{This document is the results of the BAT4ME research project funded by the Basque Country.}

\author[1]{Begoña Ispizua}[orcid = 0000-0002-6789-7866]
\ead{begona.ispizua@tecnalia.com}
\affiliation[1]{organization={Tecnalia - Basque Research Technology Alliance (BRTA)},
                 city={Derio},
                postcode={48160}, 
                country={Spain}}
\cormark[1]

\credit{Methodology, Investigation}

\author[1]{Sergio Gil-López}
\ead{sergio.gil@tecnalia.com}
\credit{Conceptualization}

\author[2]{Leire Arrizabalaga}
\ead{03leirearrizabalaga@gmail.com}
\affiliation[2]{organization={Unibersity of the Basque Country (EHU)},
                 city={Leioa},
                postcode={48940}, 
                country={Spain}}
\credit{Software}

\author[3]{Ibai Laña}[orcid= 0000-0002-2682-6199]
\ead{ilana@mondragon.edu}

\affiliation[3]{organization={Mondragon Unibertsitatea},
                 city={Bilbao},
                postcode={48001}, 
                country={Spain}}
\credit{Supervision, Writing - review and editing}
\cortext[1]{Corresponding author.}


\begin{abstract}
With the increasing integration of renewable energy sources, energy storage systems have become essential, making the accurate estimation of their State of Health (SOH) and degradation behavior critical. In this work, we propose a physics-informed deep learning approach for lithium-ion battery SOH prediction using incomplete discharge curves extracted from arbitrary voltage ranges, thereby reflecting realistic and heterogeneous operating conditions. The proposed method combines data-driven learning with physically motivated degradation dynamics to ensure consistent and reliable SOH estimation from partial discharge information, achieving a MAPE below 4$\%$. In addition, a real-time degradation trend estimation strategy is introduced to detect key aging transitions without requiring prior knowledge or historical data, making it applicable to a wide range of batteries. Overall, our approach enables SOH estimation from arbitrary discharge segments and a real-time degradation forecast that continuously integrates all usage, overcoming previous methods that rely on fixed protocols or early, non-adaptive predictions.

\end{abstract}

\begin{keywords}
Battery\sep State of Health \sep Degradation-trend \sep Prognosis
\end{keywords}

\maketitle

\section{Introduction}\label{introduction}
In recent years, Energy Storage Systems (ESS) have become indispensable. The global transition from fossil-fuel-based energy to renewable sources and electric mobility has brought along a significant increase in battery usage, as well as the development of more advanced energy storage technologies. Battery degradation is a key aspect in the management of these systems, particularly for accurate Remaining Useful Life (RUL) estimation within Battery Management Systems (BMS). This has driven the research community to focus on developing methods and techniques to improve battery lifespan.

Over the past decade, numerous studies have addressed this problem from different perspectives, which can be broadly classified into physics-based models and data-driven models. Among physics-based approaches\textemdash  methods that explicitly incorporate the underlying physics and domain knowledge governing the system\textemdash the so-called pseudo-two dimensional (P2D) model, corresponding to the spatial representation of the Doyle–Fuller–Newman model, has been widely adopted. For instance, \cite{SORDI2025100410} employs a P2D framework to diagnose lithium-ion battery degradation under automotive-like cycling conditions. To alleviate the computational burden associated with these models, reduced-order formulations have also been developed. A widely used example is the Single Particle Model (SPM), which represents each electrode as a single spherical particle and can be derived as a simplification of the P2D model \cite{batteries6030037}. In contrast, Equivalent Circuit Models (ECMs) describe battery dynamics using passive electrical components. Owing to their simplicity, they are widely employed for SOH estimation \cite{ECM_p}, however, they require accurate parameter identification and generally do not support real-time parameter updates. Beyond these modeling approaches, some studies focus on specific degradation mechanisms, such as Solid Electrolyte Interphase (SEI) degradation\textemdash the formation and evolution of the passivation layer on the surface of active anode particles\textemdash and mechanical damage at the anode particle level \cite{Karger2024}. These models offer strong physical interpretability and are generalizable without requiring training data. However, they are computationally expensive, which limits their applicability for real-time estimation.

In contrast, data-driven methods have emerged as an alternative due to their lower computational costs and fast response times. For instance, \cite{FERMINCUETO2020100006} employs a Relevant Vector Machine (RVM) to predict the degradation knee-point, a phenomenon whose importance in the degradation process is highlighted in \cite{Attia2022Knees}. Likewise, \cite{9520291} proposes a Gaussian Process Regression (GPR) model to estimate battery health changes, from which the entire capacity fade trajectory, the knee-point, and End of Life (EoL) can be inferred. Furthermore, numerous studies have demonstrated that Recurrent Neural Networks (RNNs)\textemdash particularly Long Short-Term Memory (LSTM) architectures\textemdash are well suited for capturing temporal dependencies underlying battery degradation dynamics. Based on this foundation, \cite{LI2019510} presents a hybrid approach that first decomposes capacity-per-cycle data into multiple components and then uses LSTM and Elman networks\textemdash defined by \cite{Fetanat2023FENN}\textemdash to model the low‑ and high‑frequency components, respectively. Finally, \cite{9137406} employs a Convolutional Neural Network (CNN)–LSTM framework for RUL estimation, further highlighting the effectiveness of deep temporal–spatial feature learning in this domain.

However, despite the strong experimental performance, most of the aforementioned approaches require large amounts of data, complete charge-discharge cycles, or alternatively, equal charge-discharge data segments. These assumptions considerably limit their applicability in practical scenarios, where only arbitrary portions of a charge or discharge cycle may be available. Moreover, purely data-driven methods often struggle to generalize beyond the training distribution, as they face difficulties in predicting unobserved conditions, leading to results that are difficult to interpret. 

To overcome these limitations, hybrid approaches\textemdash referred to as physics-aware machine learning (PaML) models and summarized in \cite{Xu2024PhysicsAwareML}\textemdash have emerged, combining data‑driven learning with explicit physical knowledge. For instance, \cite{Long2022AutoIP} presents Automatically Incorporating Physics (AIP), a framework capable of integrating various types of differential equations into Gaussian Processes (GPs). Similarly, \cite{Wang2021PIDeepONet} introduces physics-informed DeepONets, an extension of Deep Operator Networks (DeepONets), which have demonstrated the ability to approximate nonlinear operators between spaces. This approach incorporates physical constraints through a regularization mechanism that biases the model outputs toward physically consistent solutions. 

Among these methods, Physics-Informed Neural Networks (PINNs), introduced by \cite{RAISSI2019686}, represent one of the most prominent frameworks within the physics-aware learning paradigm. These networks are characterized by their hybrid nature, integrating data-driven learning with physics-based modeling. In particular, PINNs are trained using observational data while simultaneously incorporating the differential equations governing the underlying physical system into the loss function. As a result, the learning process is constrained to satisfy known physical laws, guiding model convergence toward physically consistent solutions and improving generalization capability, particularly in scenarios with limited data. 

PINNs have already been applied in the context of BMS. For example, \cite{Wang2024PINN} proposed a degradation-aware PINN framework in which explicit degradation equations are not available; instead, the model incorporates the prior knowledge that battery degradation implies a monotonic decay of the SOH, enforcing that successive SOH estimates decrease over time. Similarly, \cite{TANG2025825} embedded simplified voltage dynamics into the learning process to forecast future battery charging curves using data from only a single charge cycle.

Thus, PINNs provide a middle ground that can achieve better results than approaches based exclusively on either physics-based or data-driven methods. At the same time, this approach introduces challenges associated with both paradigms. On the one hand, explicit differential equations describing the system are required, which are not always accessible in practice. On the other hand, when the governing equations contain unknown parameters, observational data are still necessary for their identification. 

To address this limitation, some studies have adopted empirical or semi-empirical formulations to approximate the governing equations of different dynamic systems. For example, in \cite{CHEN2024110471}, the Arrhenius equation is used to reproduce temperature-dependent dynamics of chemical reactions, although it does not represent the full underlying physical mechanism. Similarly, semi-empirical formulations such as those proposed in \cite{Maafi2025} have been used to model kinetics of photochemical reactions.

These formulations are designed to reproduce the observed system behavior and are assumed to represent the underlying dynamics based on empirical validation. A widely used example in lithium-ion battery degradation modeling is the Verhulst equation, introduced in \cite{6587560}, and adopted in several works, including \cite{10251604, 10927632}. Nonetheless, these approaches typically rely on equal data segments under stationary charging or discharging conditions\textemdash such as a complete charge curve, a complete discharge curve, or a fixed portion of either\textemdash when predicting the SOH. This requirement is seldom met in real-world applications. For example, \cite{SUN2025126425} attempts to address this gap by proposing an approach capable of estimating the degradation trajectory at any stage of battery life, extracting this information during charging using a CNN.

From the estimated SOH, the battery degradation trajectory can be characterized, enabling the identification of key degradation events such as the degradation knee-point, where the rate of capacity loss accelerates. In contrast, the RUL represents the number of cycles remaining until the battery reaches its EoL. In \cite{9454160}, different methods for predicting both SOH and RUL are compared. In particular, it has been observed that various perspectives have been adopted for RUL prediction. For example, \cite{SUN2021108679} employs an Unscented Particle Filter (UPF) with Optimized Multiple Kernel Relevance Vector Machine (OMKRVM) for RUL estimation, while \cite{en13092380} uses a LSTM-Sliding Time Window (STW). Although these methods can provide accurate RUL estimates, they are mainly designed for early-life prediction, relying on measurements acquired during the initial battery cycles to forecast the remaining cycles until EoL. As a result, their predictions are established from limited degradation information and depend on the assumption that early-life degradation patterns remain representative throughout the battery lifetime. 

In essence, existing SOH and RUL prediction methods are built upon assumptions that limit their applicability in practical scenarios, such as the availability of complete charge/discharge curves or the representativeness of early-life degradation patterns. Although some studies have demonstrated that the entire discharge curve is not required for SOH prediction, such as \cite{batteries11050167}, they still rely on the same predefined charge/discharge region across all batteries. This assumption overlooks the fact that different voltage regions may exhibit different degradation sensitivities and, therefore, require different predictive models and features. To overcome these limitations, the present work proposes a more generalizable and practical approach that does not depend on complete charge–discharge curves for SOH estimation or on early-life data for reliable RUL prediction.

In this work, we propose a physics-informed deep learning framework for lithium-ion battery SOH estimation from arbitrary charge or discharge segments by automatically identifying the corresponding voltage region and selecting a dedicated prediction model. As a preliminary step, the study investigates whether charge or discharge data provide more informative inputs for SOH estimation, since this choice determines the subsequent analyses and model configuration. Unlike conventional approaches, the proposed method neither requires complete charge/discharge curves nor relies on predefined voltage windows, enabling SOH estimation from arbitrary charge or discharge segments. This flexibility enables the use of arbitrary input information and enhances the applicability of the method under real-world operating conditions. The framework incorporates a set of features selected and validated through machine learning techniques for each charge or discharge segment, rather than assuming a universal feature set applicable to all segments. By embedding physical constraints within the learning process, the proposed PINN achieves accurate SOH estimations. To assess the contribution of the physics-informed formulation, its performance is systematically compared with that of a conventional feedforward neural network (FNN), highlighting the advantages of incorporating physical knowledge into the learning process.

Finally, a real-time degradation prognosis framework is proposed, operating on the available SOH estimates and continuously updating the predicted degradation trajectory as new measurements become available. The framework performs the prognosis independently for each battery cell, without requiring a predefined degradation model or battery-specific parameters. Based on the evolving degradation trajectory, an adaptive linear regression scheme identifies both the degradation knee-point\textemdash the transition from approximately linear to accelerated capacity fade\textemdash and the EoL.

The experimental setup and the results obtained allow the following four Research Questions (RQs) to be addressed:
\begin{itemize}
\item \textbf{RQ1:} Which process\textemdash charge or discharge\textemdash provides more informative data for battery SOH estimation?
\item \textbf{RQ2:} Can battery SOH be accurately estimated from an arbitrary charge or discharge segment, without requiring complete curves or predefined voltage regions?
\item \textbf{RQ3:} Do different voltage regions require distinct predictive features for accurate SOH estimation?
\item \textbf{RQ4:} Can the battery degradation trajectory be reliably estimated online from continuously updated SOH predictions, without requiring prior battery-specific information or predefined degradation models?
\end{itemize}




The article is organized as follows. Section \ref{material_method} describes the data used and the methodology followed. It is structured into three subsections: Subsection \ref{data_processing} covers the data processing procedure, Subsection \ref{PINN} presents the methodology used to define the PINN, and Subsection \ref{lse} details the procedure for defining the real-time degradation estimation. Section \ref{results} reports the experimental results and finally, Section \ref{conclusions} summarizes the main conclusions of this work.

\section{Materials and Methods}\label{material_method}
\begin{figure}[ht]
    \centering
    \includegraphics[width=\columnwidth]{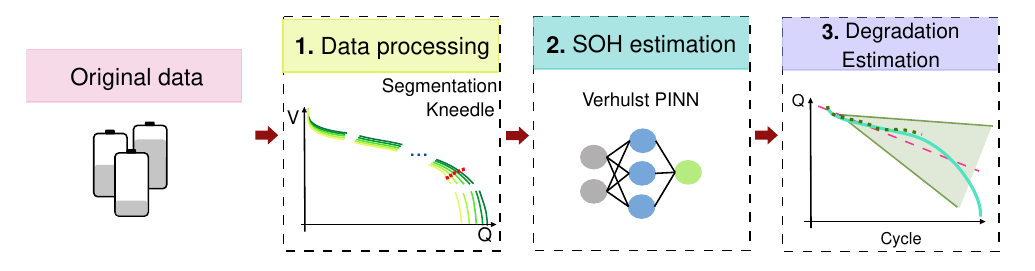}
    \caption{Overall workflow of the proposed methodology}
    \label{fig:diagram-scheme}
\end{figure}

This section presents the methodology used to: first, process the data to obtain different sections of each curve and extract the corresponding features; then, design the NN, based on the extracted features to predict the SOH; and finally, estimate the degradation tendency based on the approached SOH. The methodology is organized into three main stages, represented in Figure \ref{fig:diagram-scheme}, each addressed in a dedicated subsection.

The dataset used in this study, presented in \cite{Severson2019}, comprises 122 lithium-ion batteries cycled until their End of Life (EoL), each with a nominal capacity of 1.1Ah and, a nominal voltage of 3.3V,  with an upper cut‑off voltage of 3.6V and a lower cut‑off voltage of 2V. All cells are discharged using a constant current of 4C, while different charging protocols are applied. Specifically, the dataset includes both one-step and two-step fast-charging strategies, which have been shown to significantly influence battery degradation behavior \cite{Xu2021}. In the two-step protocol, a constant current is applied until the battery reaches 80\% of its State of Charge (SOC), after which, a different current is used to complete the charging process. The dataset provides time-resolved measurements of voltage, current, and capacity.

\subsection{Data processing}\label{data_processing}

As mentioned previously, knowledge of the battery SOH is critical for BMS and, consequently, for improving battery lifespan. However, this value is not always accessible, particularly when the charge or discharge curve is incomplete. Although SOH can also be obtained through electrochemical characterization methods \cite{LI20181178}, such approaches are not feasible in real operating conditions. Therefore, one of the main challenges addressed in this work is the estimation of SOH from an incomplete charge or discharge curve using a PINN, which integrates both data-driven information and physical principles. 

\begin{figure}[h!]
    \centering
    \includegraphics[width=0.8\columnwidth]{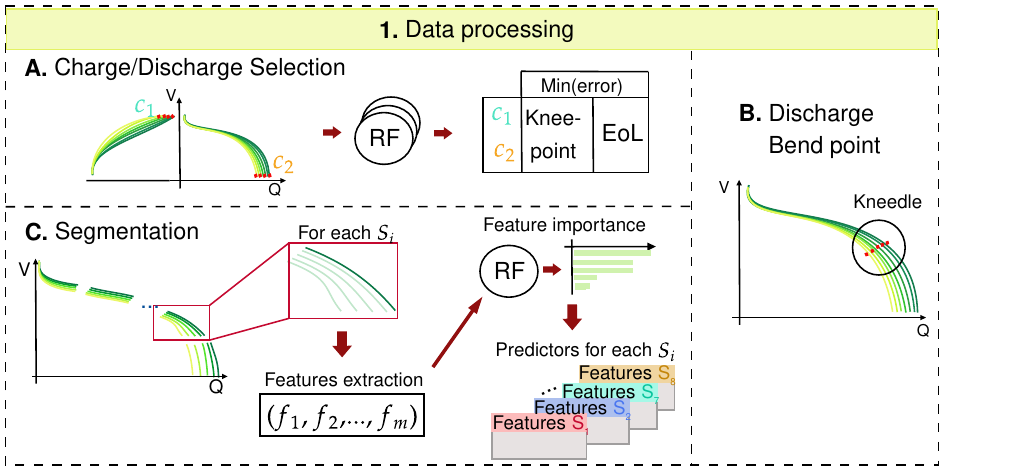}
    \caption{Workflow of the data processing section}
    \label{fig:data_processing_plot}
\end{figure}

In Figure \ref{fig:data_processing_plot}, the subsections within the data processing workflow are illustrated.

\subsubsection{Charge/Discharge curve selection}\label{curve_selection}

The first step, represented in Figure \ref{fig:data_processing_plot} as \textbf{A}, is to determine which process\textemdash charge or discharge\textemdash provides the most informative data for training the PINN. Previous studies have employed either charging or discharging data for SOH estimation. Charging data are often preferred due to their higher consistency and stability, e.g., \cite{PETRELLA2025118747}, whereas discharge data are attractive because they reflect battery behavior under real load conditions, e.g., \cite{YANG2022103857}. However, there is no consensus in the literature regarding which phase contains more informative degradation signatures for SOH prediction. To address this gap, this work performs a comparative analysis of degradation-related predictions\textemdash specifically, the prediction of the degradation knee-point, i.e., the point at which the degradation regime changes, and EoL\textemdash using both charge and discharge data, following the methodology proposed in \cite{Severson2019}. In that study, the SOH is defined as the capacity measured when the battery is fully discharged, with the capacity given by
\begin{equation} \label{eq:capacity}
    Q = \int _t I(t) \cdot dt,
\end{equation} where $I(t)$ represents the discharge current.

The first 100 cycles of the degradation curve, corresponding to the initial stages of degradation, are used to perform these predictions, as in the aforementioned study.

Three different models are considered to ensure the robustness of the comparison: Random Forest (RF) \cite{Breiman2001}, Extreme Gradient Boosting (XGB) \cite{Chen2016XGBoost}, and an Feedforward Neural Network \cite{Rumelhart1986Backprop}. Their performance is evaluated using the Mean Absolute Percentage Error (MAPE). Based on the results of this comparison, the remainder of the study focuses on the most informative phase for SOH estimation, namely the one that provides the highest predictive accuracy. This process strengthens the robustness of the method by providing empirical justification for the data used.

\subsubsection{Discharge bend point and computation of the SOH}\label{output}
Represented in Figure \ref{fig:data_processing_plot} as \textbf{B}, this subsection describes the procedure used to identify the bend point of the discharge curve and its relationship with battery SOH. 

To define a PINN capable of predicting battery SOH, a reliable ground truth for model training and validation must first be established. Accordingly, the SOH can be defined using the widely adopted formulation reported in \cite{S2024123542} as
\begin{equation}
    SOH = \frac{Q_{act}}{Q_{nom}}\cdot 100,
\end{equation}
where $Q_{act}$ and $Q_{nom}$ are the actual and nominal capacities, respectively.

However, \cite{Severson2019} defines the health indicator as the discharge capacity measured at the cut-off voltage of each cycle, which is then used to construct the degradation curve. Building on this idea, \cite{begoIbai} proposed a novel indicator based on the capacity at the bend point of the discharge curve, arguing that it is more representative of battery aging. Moreover, this capacity can be obtained before the battery is fully discharged, making it more suitable for real-world applications. Since the bend point occurs at approximately the same voltage under a constant discharge current, the corresponding capacity is directly comparable across cycles. Furthermore, because the dataset consists of complete cycles with a constant depth of discharge (DoD), the evolution of this indicator can be represented directly as a function of the cycle number. If, however, the cycles reached different DoD levels, the measured capacity would need to be normalized with respect to the corresponding depth of discharge to ensure comparability across cycles.

Accordingly, the capacity at the bend point is adopted in this work as the health indicator and is hereafter referred to as SOH, although it does not correspond to its conventional definition. Nevertheless, the proposed methodology is not restricted to this particular indicator and can be readily extended to any other SOH definition, provided that it is used consistently throughout the training and evaluation stages.

These points are identified using the Kneedle algorithm, which detects the point of maximum curvature by finding the greatest perpendicular distance between the curve and the line connecting its endpoints, as described in \cite{5961514}.

The resulting discharge bend point based SOH values are computed for each cycle of each battery, and subsequently used as the ground truth outputs for training and validation of the proposed PINN models, as they define the corresponding degradation curves.

\subsubsection{Data segmentation and features extraction}\label{input} 
Figure \ref{fig:data_processing_plot}\textbf{C} illustrates an overview of the main processing steps of the proposed approach. As previously stated, one of the key innovations of this study is the estimation of the battery SOH using only limited portions of the discharge process, rather than requiring complete discharge curves. This approach is applicable across different voltage regions of the discharge curve, enabling SOH estimation from arbitrary discharge segments.

To this end, each discharge curve is segmented into 8 sections, uniformly distributed between the upper and lower cut-off voltages. Each segment corresponds to a fixed capacity decrement of 5\% relative to the nominal capacity. These parameters were selected as a balance between increasing the number of segments per discharge curve and maintaining sufficient segment length. Increasing the number of segments would lead to a significant overlap between them, whereas reducing their length would result in subcurves that are too short to reliably extract features.

Once the discharge curves are segmented, it is worth noting that, unlike the original dataset\textemdash where cumulative discharge capacity is available\textemdash{}, this information is no longer accessible for individual segments. This scenario is consistent with real-world operating conditions, in which only local measurements over limited time intervals are typically available. Therefore, the capacity associated with each discharge segment is computed independently according to the definition given in Equation \ref{eq:capacity}. 

All subsequent processing steps are applied independently to each discharge segment, which is treated as an individual sample. This approach relies on the assumption that the best SOH-predicting features are not necessarily the same across different discharge segments, and they may therefore vary depending on the voltage range of the discharge process.

This design enables the model to consistently handle discharge intervals extracted from any region of the complete discharge process. Once trained, the model identifies the voltage range of a newly acquired discharge record and selects the corresponding model for SOH estimation, extracting the required features accordingly.

Based on the assumption that relevant information is captured by different features in each segment, a segment-specific set of features is extracted for SOH estimation. While the relevance of individual features may vary across discharge segments, an initial common feature-selection procedure is applied to all segments to ensure methodological consistency and comparability.

Since a voltage–capacity (VQ) curve can be defined for each segment, a set of geometric descriptors is computed, including arc length, slope, and enclosed area. In addition, statistical indicators\textemdash mean, variance, kurtosis, and skewness of both current and voltage\textemdash are extracted. Finally, features derived from the incremental capacity (IC) curve are incorporated, which has been demonstrated to be strongly related to SOH in \cite{8603757}. These IC-based features include the enclosed area, the maximum value of the $dQ/dV$ ratio, and the mean value of its second derivative.

\begin{table}[H]
\caption{Extracted features from the voltage based segment.}
\centering\label{tab:features}
\begin{tabular}{lcl}
\toprule
\textbf{Feature group} & \textbf{Symbol} & \textbf{Description} \\ \midrule
\multirow{3}{*}{VQ curve} 
    & $L_{VQ}$ & Length of the arc \\ 
    & $m_{VQ}$ & Slope \\ 
    & $A_{VQ}$ & Enclosed area \\ 
\midrule
\multirow{4}{*}{Voltage} 
    & $\mu_V$ & Mean of the section \\ 
    & $\sigma_V^2$ & Variance of the section \\ 
    & $\kappa_V$ & Kurtosis of the section \\ 
    & $\gamma_V$ & Skewness of the section \\ 
\midrule

\multirow{4}{*}{Current} 
    & $\mu_I$ & Mean of the section \\ 
    & $\sigma_I^2$ & Variance of the section \\ 
    & $\kappa_I$ & Kurtosis of the section \\ 
    & $\gamma_I$ & Skewness of the section \\ 
\midrule
\multirow{3}{*}{IC curve}
    & $A_{IC}$ & Enclosed area \\
    & $max_{IC}$ & Maximum value \\
    & $\mu_{IC'}$ & Mean of the derivative \\ 
\bottomrule
\end{tabular}
\end{table}

Accurate SOH estimation critically depends on the selection of appropriate predictors. In conventional statistical analyses, relevant features are commonly identified using correlation-based tests \textemdash such as Pearson's correlation for linear relationships\textemdash along with significance tests to account for potential nonlinear effects. However, \cite{RF_featu_importance} demonstrated that feature importance ($FI$) metrics derived from RF models provide a robust assessment of the relative contribution of each predictor to the output. 

Within the RF framework, each feature is assigned an importance score, normalized such that the importance values of all features sum to one. Therefore, if all $|N|$ features contributed equally to the prediction, each would be expected to have an importance of $FI = 1/|N|$. This value provides a natural baseline for feature selection: predictors with importance values exceeding $1/|N|$ are considered to contribute more than expected under an equal-contribution assumption and are therefore retained for SOH estimation.

Thus, following this feature selection stage, 8 different datasets are obtained\textemdash one for each discharge segment\textemdash each characterized by its own subset of significant features. These datasets are then used to train separate NN models.

Following the segmentation stage, each subset is treated independently. Consequently, each voltage-based segment has its own feature selection process and a dedicated NN model for SOH estimation. As a first step in the feature selection process, a common set of candidate features is defined for all discharge segments to ensure methodological consistency. This preselection, summarized in Table \ref{tab:features}, comprises a total of 15 features.

In addition, to make the proposed framework applicable to arbitrary discharge segments, an XGBoost classifier is trained to identify the corresponding voltage-range category. The classifier uses the features listed in Table \ref{tab:features}, excluding the voltage statistics $\mu_V, \sigma_V^2, \kappa_V, \gamma_V$ to reduce the risk of overfitting. Once a discharge segment has been classified, the predicted category is used to automatically select the corresponding NN model for SOH estimation. Therefore, given an arbitrary discharge segment of any length, the predefined features are extracted and provided to the XGBoost classifier, which assigns the segment to one of the eight voltage-range categories and, consequently, selects the appropriate NN model.

\subsection{SOH estimation} 
\label{PINN}
Once the data have been processed, as depicted in Figure \ref{fig:diagram-scheme}, the next step is to define a NN capable of accurately estimating the battery SOH, using as input the segment-level features extracted in the previous section. This is addressed using a Verhulst PINN.

Within the context of BMS, explicit degradation-related differential equations are generally unavailable, which limits the direct application of PINNs. In this context, the Verhulst equation \cite{Bacaer2011} emerges as a suitable approximation of lithium-ion battery degradation curves. The Verhulst model is defined as a logistic differential equation:
\begin{equation}\label{eq:verhuls_0}
\begin{split}
    \frac{du(t)}{dt} &= ru(t)  \left[ 1-\frac{u(t)}{K} \right ], \\
    &\text{ s.t. } r>0, \text{ and } 0<u(t)<K, \\
\end{split} 
\end{equation}
where $u(t)$ is the capacity loss at time $t$ (that is, $u(t) = 1 - SOH(t)$), $r$ is the degradation constant governing the rate of degradation and $K$ is the upper bound of capacity loss, assuming that the degradation is finite. Both $r$ and $K$ must be estimated. In this work, the time variable $t$ corresponds to the cycle number, considering a cycle as a full charge-discharge process.

The PINN formulation adopted in this work follows the methodology proposed in \cite{10251604}. Its main components are briefly summarized below.

However, modeling SOH solely as a univariate function of time is insufficient to distinguish degradation trajectories across different batteries. To address this limitation, latent variables acting as health indicators are introduced, represented by $\vec{x} = [x_1, \dots, x_n] \in \mathbb{R}^n$. This formulation enables the incorporation of monitoring data and features as a multidimensional health indicator. Furthermore, assuming an initial degradation occurring before the battery is put into use \textemdash a degradation not captured by Equation \ref{eq:verhuls_0}\textemdash an additional constant $C$ is introduced to represent this initial capacity loss. The resulting generalized Verhulst model is expressed as:
\begin{equation}\label{eq:Verhulst}
\begin{split}
    \frac{\partial u(\vec{x}, t)}{\partial t} &= r[u(\vec{x},t) -C ] \left[ 1-\frac{u(\vec{x},t) -C }{K-C} \right] , \\
    &\text{ s.t. } r>0,\\
    &0<u(\vec{x}, t)<K, \\
    &\text{ and }0<C\leq u_0.\\
\end{split}
\end{equation}
According to \cite{10251604}, $u_0$ is commonly set to $10\%$ and $K\in [20\%,100\%]$.

Following this formulation, a surrogate NN is embedded within a PINN formulation by enforcing consistency with the Verhulst degradation dynamics. The network output $u(\vec{x}, t)$ represents the SOH, and its temporal derivative 
is computed via automatic differentiation with respect to the cycle index. Based on this formulation, the physical residual is then defined as 
\begin{equation} \label{eq:l_f}
\mathcal{L}_f =\frac{\partial u(\vec{x}, t)}{\partial t} - G(u(\vec{x}, t);r,K,C), \end{equation}
where $G(\cdot)$ denotes the Verhulst model (Equation \ref{eq:Verhulst}), parameterized by $r$, $K$, and $C$, which are unknown and are also optimized during training. In this study, as in \cite{10251604}, since all batteries are identical, these parameters are assumed to be identical across all cells. In more heterogeneous settings, battery-specific parameter optimization would be required.

In PINNs, physical constraints are most commonly incorporated through the loss function. While, in a conventional FNN, the loss function is defined solely by the discrepancy between the observed output and the network predictions, physics-informed models augment this formulation by introducing an additional term that enforces consistency with known governing physical equations. Hence, the training objective is formulated by combining two complementary loss components: a data-driven loss and a physics-based loss. The data loss, denoted as $\mathcal{L}_u$, measures the discrepancy between the network predictions and the available ground-truth SOH values. The physics loss, $\mathcal{L}_f$, penalizes violations of the governing degradation dynamics in Equation \ref{eq:l_f}. By jointly minimizing these two terms, the learned SOH trajectories are constrained to remain both accurate with respect to the data and consistent with physically plausible degradation behavior.  

Accordingly, the total loss function of a PINN can be expressed as 
\begin{equation}\label{eq:PINN_loss} 
\mathcal{L}_{\mathrm{PINN}} = \lambda_u \mathcal{L}_u + \lambda_f \mathcal{L}_f , \end{equation}
where $\lambda_u$ and $\lambda_f$ control the relative importance of data and physical consistency during training and must be optimized. 

The data-driven loss, $\mathcal{L}_u$, is defined as the Mean Squared Error (MSE), which is widely adopted due to its strong penalization of large prediction errors during training. For validation, the MAPE, the Root Mean Squared Error (RMSE), and the coefficient of determination ($R^2$) are employed to assess the prediction accuracy.

To quantify the contribution of the physics-informed formulation, the proposed PINN is also compared with a conventional FNN trained under the same experimental conditions. This comparative analysis validates the benefit of incorporating physical constraints into the learning process.

This framework provides accurate SOH predictions, a parameter that is essential for BMSs as they face the challenge of optimizing the battery lifespan. In this work, the analysis goes a step further by using this predicted value to estimate the battery degradation, i.e., the decay of SOH. Once the PINN is trained, it delivers SOH estimates instantaneously, allowing this information to be incorporated in real time for a more precise degradation assessment.

\subsection{Degradation estimation} \label{lse}

\begin{figure}[h!]
    \centering
    \includegraphics[width=0.7\columnwidth]{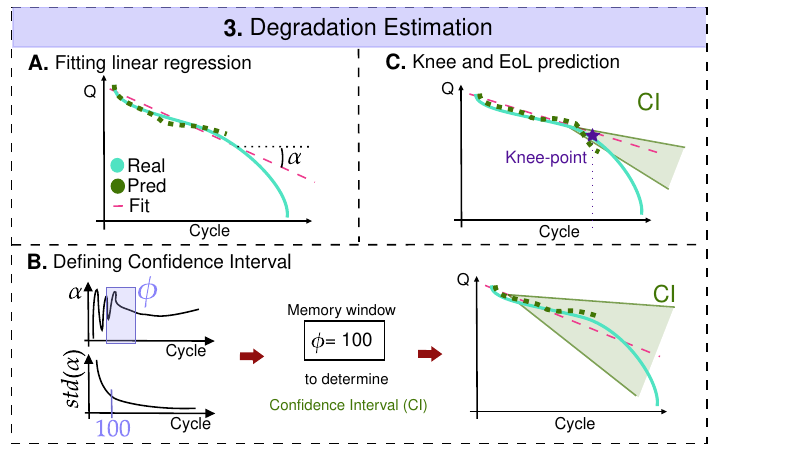}
    \caption{Workflow of the degradation estimation process}
    \label{fig:eskema_degradation}
\end{figure}

Once the SOH is estimated, a real-time prediction of the battery degradation trend is proposed. Figure \ref{fig:eskema_degradation} illustrates the process developed to address this part of the study\textemdash the third and final stage\textemdash as represented in Figure \ref{fig:diagram-scheme}.

This sections is organized into three subsections. First, the motivation for fitting the degradation curve by a linear regression is analyzed. Then, the definition of the confidence interval for the fitted line is examined. Finally, the identification of the knee-point of the degradation curve is addressed.

\subsubsection{Fitting a linear regression}\label{fittin_linear_regression}

Numerous machine learning and deep learning approaches have been proposed to forecast either the number of cycles remaining until the knee-point or until EoL. These approaches typically require large amounts of similar historical data from batteries with known degradation trajectories. Among degradation modeling efforts, significant attention has been dedicated to the early prediction of degradation features, such as the knee-point or the cycle life. However, these approaches do not consider changes in usage behavior. As a result, if operating conditions change after the early prediction, the forecast may not be valid, since such changes are not incorporated into the model. In contrast to these methodologies, this work adopts a computationally efficient strategy based on a linear least-squares approximation, represented by block  as \textbf{A} in Figure \ref{fig:eskema_degradation}. 

\begin{figure}[h!]
    \centering
    \includegraphics[width=0.5\columnwidth]{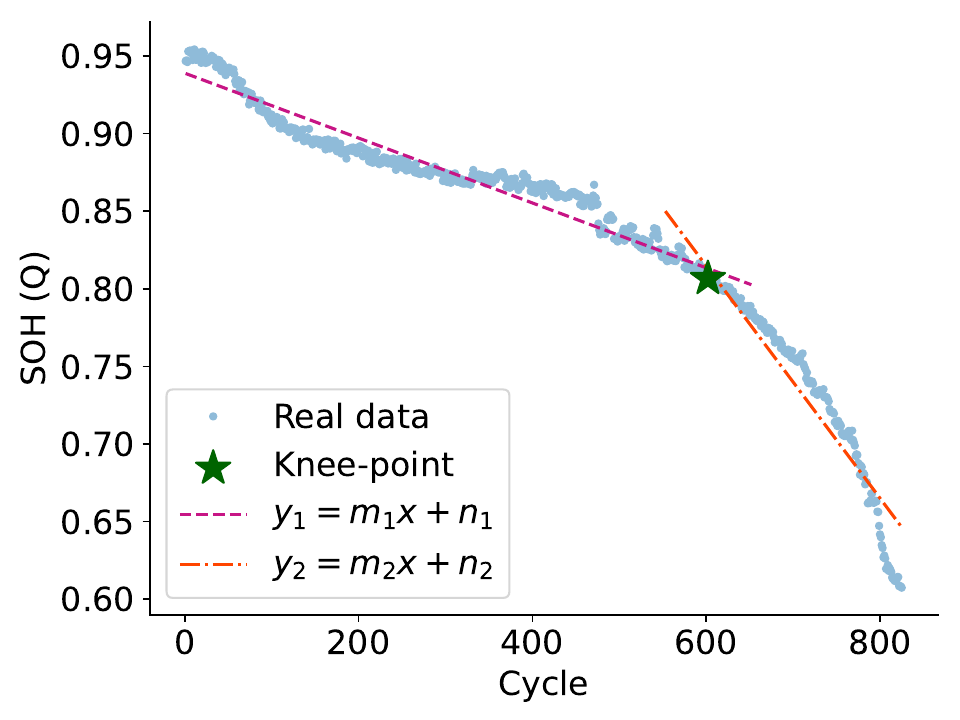}
    \caption{Linear regression approaches of degradation trend}
    \label{fig:linear_approach}
\end{figure}

After analyzing various degradation curves, such as those presented in \cite{Severson2019}, and considering Equation \ref{eq:verhuls_0}, where $r$ represents the degradation constant, this section proposes modeling the degradation trend using a linear regression up to the knee-point\textemdash the point at which the degradation trend changes and a new regime can be identified. Figure \ref{fig:linear_approach} illustrates these two regimes that intersect at the knee-point.

The horizontal axis corresponds to time, expressed in number of cycles, while the vertical axis shows the SOH. The degradation curve is approximated by two linear regimes\textemdash pink and orange\textemdash that intersect at the knee-point, represented by a star. The first regime, $y_1$, is characterized by a slope $m_1$ and an intercept $n_1$, while the second, $y_2$, has a slope $m_2$ and an intercept $n_2$, defined at $x=0$.

However, the purple line in Figure \ref{fig:linear_approach} is obtained, for illustration purposes, by fitting all SOH values predicted up to the knee-point. In practice, the main interest lies in defining a real-time degradation trend\textemdash that is, updating the regression whenever a new discharge process is completed, and consequently, as a new SOH value becomes available. Therefore, the regression is drawn starting from the second prediction onward and is continuously updated whenever a new SOH estimate is obtained.

\subsubsection{Defining a Confidence Interval}\label{confidence_interval}

As shown in Figure \ref{fig:linear_approach}, the first linear segment\textemdash in purple\textemdash is employed to capture the degradation trend prior to the beginning of accelerated aging and the corresponding knee-point. However, in an online setting, the knee-point is not known a priori, therefore, it is not possible to determine in advance when the decay transitions to a new degradation regime. Instead, as the SOH is estimated at each cycle, the parameters of the degradation line are continuously updated using a least-squares fit. Since a straight line is fully defined by two points, each newly estimated SOH value contributes to refining the degradation trend. During the early cycles, the slope of the fitted line may vary substantially due to the limited amount of data, but these variations decrease as more cycles are incorporated. From a certain point onward, the degradation trend becomes sufficiently stable, enabling reliable real-time tracking.

\begin{figure}[h!]
    \centering
    \includegraphics[width=1\columnwidth]{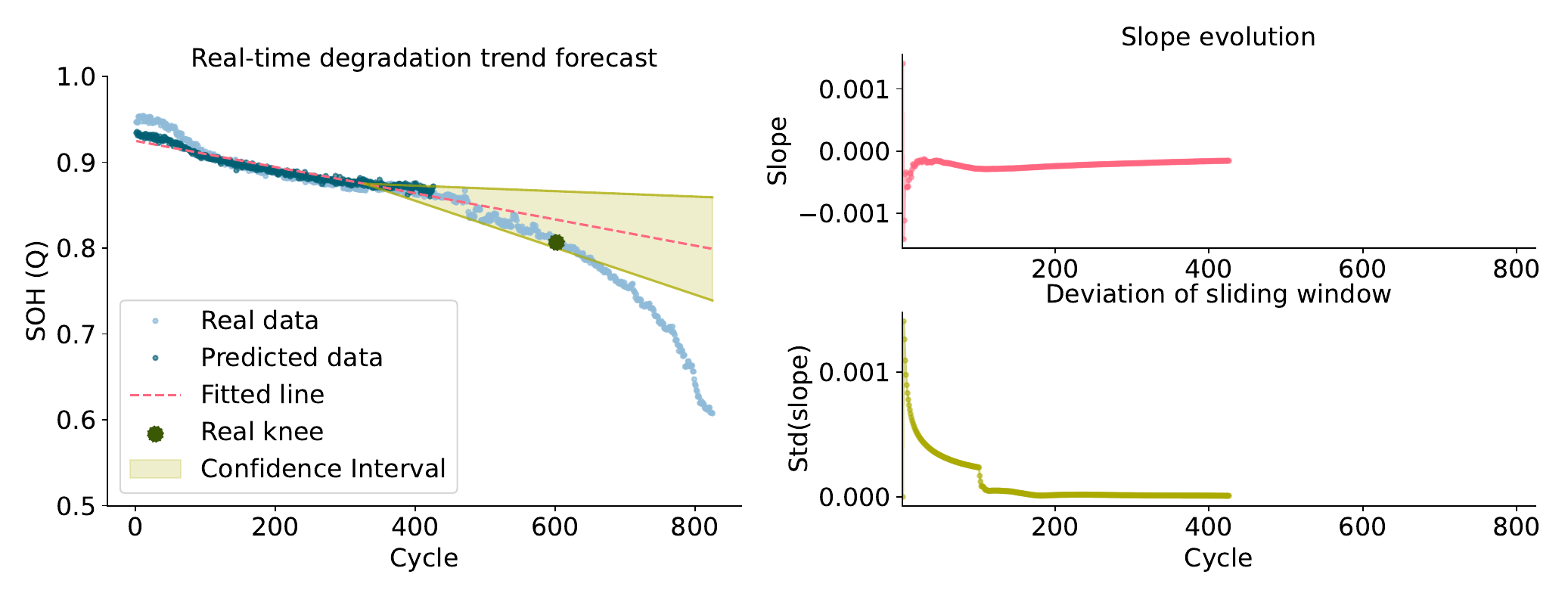}
    \caption{Degradation trend estimation and stability analysis. Left: fitted degradation line (dashed pink) with its corresponding confidence interval (shaded green), predicted SOH values (green), and the true knee-point indicated by dark green star-shaped marker. Right: temporal evolution of the slope (top) and the standard deviation of the slope computed over a sliding window of 100 cycles (bottom).}
    \label{fig:trend}
\end{figure}

The key aspect of this approach is the identification of the cycle at which the degradation line becomes stable and can be assumed to represent the underlying degradation trend. This process is identified in Figure \ref{fig:eskema_degradation} as \textbf{B}. To determine this cycle, the temporal evolution of the slope and the deviation of its successive variations are analyzed, illustrated in Figure \ref{fig:trend}. Specifically, the slope is tracked as a function of cycle number, and the sstandard deviation of slope variations up to each cycle is computed. 

This deviation is expected to follow a decaying trend, similar to an $e^{-t}$ function, as the magnitude of slope variations diminishes with the increasing amount of available data. Beyond a certain point, this decay rate changes, indicating that the influence of incorporating a new slope estimate on the standard deviation calculation becomes smaller than that of adding a new SOH observation. 

Based on this observation, the deviation is computed using a sliding window \textemdash represented in Figure \ref{fig:eskema_degradation}\textbf{B} as $\phi$\textemdash{} whose length corresponds to the cycle at which the change in the standard deviation of the slope trend is identified. This deviation is updated at each cycle and is used to define the Confidence Interval (CI) of the fitted degradation line, which is expected to progressively narrow as more data become available, as illustrated in Figure \ref{fig:trend}. Specifically, the CI is bounded by two linear regressions that intersect the fitted degradation line at the point located $\phi$ cycles before the current time step. The slopes of these two lines are given by the slope of the fitted regression plus and minus the standard deviation $\sigma$ of the slopes estimated over the previous $\phi$ fitted regressions, thereby defining the upper and lower confidence bounds.

At this stage, the study develops a method capable of accurately estimating the SOH from a short segment of battery usage data, as in real-life operation conditions. Moreover, this health indicator is then used to predict the degradation trend that the battery is expected to follow if it continues to operate under similar conditions. However, detecting the knee-point of the degradation trajectory remains a challenge.

\subsubsection{Knee-point identification}\label{knee-point_identification}
In this section, represented graphically in Figure \ref{fig:eskema_degradation}\textbf{C}, the detection of the knee-point\textemdash and therefore the beginning of the exponential degradation of the battery until the EoL\textemdash is explored. This analysis is addressed through the identification of two alarms: the first signaling the transition to an accelerated degradation regime, and the second indicating the approach of the EoL. 

It can be assumed that once the degradation trend is fitted and its associated confidence interval is obtained, a few SOH predictions falling outside this interval may indicate a change in the degradation regime. However, a single outlier is not sufficient to reliably declare a trend change, so a criterion must be defined regarding the number of consecutive predictions outside the confidence interval required to trigger the alarm.

To address this issue, a sensitivity analysis is conducted for each validation battery by varying the alarm threshold, defined by $\gamma$ as the consecutive SOH predictions outside the confidence interval. In other words, the estimated knee-point cycle is computed when there is only one value outside the CI ($\gamma = 1$), when there are two ($\gamma = 2$), and so on. For each $\gamma$ threshold, the estimated knee-point cycle is compared with the true knee-point, and the resulting error\textemdash expressed as a percentage of the total battery lifetime, allowing comparison across batteries with different lifetimes\textemdash is defined as Knee-point Percentage Error (KPPE), which is mathematically formulated as follows
\begin{equation}\label{eq:KPPE}
\mathrm{KPPE}_{\gamma} =  \frac{\left| c_{\gamma} - c_{\mathrm{obs}} \right|}{c_{\mathrm{EoL}}} \times 100,
\end{equation}
where $c_{\gamma}$, and $c_{\mathrm{obs}}$ denote the predicted and observed cycles until the knee-point, respectively for each $\gamma \in[1,70]$, and $c_{\mathrm{EoL}}$ is the total number of cycles until EoL.

Based on the average behavior across all batteries, the optimal value of $\gamma$ is determined. Once this threshold is reached, indicating that the knee-point has been detected, an alarm is triggered, and all subsequent SOH values are considered to belong to a new degradation regime 

 
To model the new degradation regime, a second least-squares regression is applied, and its corresponding CI is determined using the procedure described in Section \ref{confidence_interval}. Although the battery lifetime threshold, defined by a minimum capacity value, can be specified by the user, it is commonly assumed that the battery reaches its EoL when its capacity drops to 80\% of its nominal capacity. At the selected minimum capacity threshold, both the fitted line and the CI intersect, and the cycle at which this intersection occurs is considered the expected EoL of the battery. However, the linear regression is continuously updated as new SOH estimates become available. Once the SOH estimations reach the defined capacity limit, a final alarm is triggered, indicating that the battery has reached its EoL.

This methodology is computationally efficient and does not require prior battery-specific information or calibration. It can be applied across different battery datasets, charging and discharging protocols, and operating voltage ranges, as it captures cycle-level degradation while continuously adapting to changes in degradation behavior. If SOH estimates are not directly available, the first stage of the proposed framework can be used to obtain them. Otherwise, the proposed real-time degradation prognosis method can be applied directly.

\section{Results}
\label{results}

This section presents the results obtained throughout the processing stages described in the previous sections. The experiments were carried out with the dataset described in Section \ref{material_method}. From the original dataset, 10 batteries were separated and exclusively used for model validation, while the remaining cells were employed for training and analysis. All results presented in this section refer to this validation set of batteries.

As a first step, a preliminary analysis was conducted to assess whether charge or discharge data provide greater predictive capability for degradation-related features, in particular for knee-point and EoL prediction, Figure \ref{data_processing}\textbf{A}. 

\begin{table}[H] 
\caption{MAPE results for knee-point and EoL prediction}
\label{tab:leire-results}
\vspace{0.1cm}
\centering
\begin{tabular}{lcccc}
\toprule
 & \multicolumn{2}{c}{\textbf{Knee-point}}&\multicolumn{2}{c}{\textbf{EoL}}\\
\cmidrule(lr){2-3} \cmidrule(lr){4-5} 
\multicolumn{1}{c}{\textbf{Model}}&Charge & Discharge &Charge & Discharge\\
\midrule
Random Forest & 38.92 & 28.67  & 35.70 & 24.56\\
XGBoost & 39.51 & 32.08 &  41.28 & 26.85\\
Neural Network & 36.58 & 29.78 & 34.07 & 26.14\\ 
\bottomrule
\end{tabular}
\end{table}

Table \ref{tab:leire-results} summarizes the prediction performance for knee-point and EoL estimation using both charge and discharge data. Discharge data consistently achieves lower MAPE values than charge data, across all evaluated models and prediction targets. This indicates that discharge profiles capture more informative degradation patterns and provide superior predictive capability for SOH-based degradation forecasting, answering \textbf{RQ1}. Based on this comparative analysis, discharge curves are selected as the basis for all subsequent experiments. 

Following the procedure presented in Figure \ref{fig:diagram-scheme} and addressing \textbf{RQ2}, the next stage is to predict the battery SOH when only a limited portion of the discharge process is available. To achieve this, the ground-truth SOH must be known for each cycle, and used as the target for the NN. Therefore, prior to dividing the original dataset into smaller subsets, the bend point of each discharge curve is identified, using the Kneedle algorithm, and the corresponding capacity values are used as SOH indicator for each cycle. 

To proceed with the study, the original dataset is first partitioned into 8 voltage-based subsets. Each selected voltage defines the endpoint of a discharge segment, which extends along the discharge curve backward until the voltage corresponding to a 5$\%$ increase of the nominal capacity (0.055 Ah) is reached. All segments are extracted in the charging direction, except for the final one at the cut-up voltage. Since the uniform discretizations include the maximum cut-up voltage (3.6 V), the last segment is constructed in the discharging direction to obtain the required capacity interval.

As an intermediate validation of the proposed pipeline, the XGBoost classifier used to identify the voltage-range category of an input discharge segment achieves an accuracy of 0.99 on the validation set. This result demonstrates that previously unseen discharge segments can be reliably assigned to the corresponding voltage-range category and, consequently, to the appropriate NN model for SOH estimation. Although the batteries considered in this study exhibit relatively homogeneous discharge characteristics, this classification stage extends the applicability of the framework to more heterogeneous battery datasets, where the voltage range of an input segment cannot be assumed a priori.

After validating the classification stage, feature selection is performed independently for each of the eight voltage-based segments using the RF feature importance scores described in Subsection \ref{input}. Table \ref{tab:feature_selection} summarizes the subset of predictors selected for each segment.

\begin{table}[H]
\caption{Selected features for each discharge segment.}
\label{tab:feature_selection}
\vspace{0.1cm}
\centering
\begin{tabular}{cl}
\toprule
\textbf{Segment}& \textbf{Selected features} \\ 
\midrule
1 &  $\mu_V$, $\kappa_V$, $A_{IC}$, $m_{VQ}$, $L_{VQ}$ \\ 
2 &  $L_{VQ}$, $A_{IC}$, $\max_{IC}$, $m_{VQ}$ \\ 
3 & $L_{VQ}$, $m_{VQ}$, $\max_{IC}$, $A_{IC}$, $\mu_{IC'}$ \\ 
4 & $m_{VQ}$, $\max_{IC}$, $A_{IC}$, $\sigma_V^2$ \\ 
5 & $L_{VQ}$, $A_{IC}$, $A_{VQ}$, $\mu_{IC'}$, $\max_{IC}$ \\ 
6 &  $m_{VQ}$ \\ 
7 &  $\gamma_I$, $A_{VQ}$, $m_{VQ}$, $A_{IC}$, $\mu_{IC'}$ \\ 
8 &  $\sigma_V^2$, $\mu_{IC'}$, $L_{VQ}$, $A_{VQ}$, $\mu_V$
\\ 
\bottomrule
\end{tabular}
\end{table}

It is worth noting that all segments are characterized by 4 or 5 significant features, with the exception of the 6th segment, for which only the slope of the VQ curve is retained. This is because segment 6 corresponds to the most linear portion of the discharge curve, located immediately before the knee-point. Table \ref{tab:feature_selection} addresses \textbf{RQ3}, showing that different voltage‑based segments require different predictive features, rather than relying on a common feature set across the entire discharge curve.

Subsequently, once the predictors have been selected for each voltage segment, a PINN is trained independently for each segment. Each model consists of two main components: a data-driven surrogate NN and a physics-based constrain derived from the Verhulst model.

The surrogate model is implemented as a fully connected feedforward neural network. Physical constrains, Equation \ref{eq:Verhulst}, are incorporated into the loss function, where the physical parameters $r$, $K$, and $C$ are treated as trainable variables and are optimized jointly with the NN weights via backpropagation. To ensure physical interpretability and numerical stability, these parameters are reparameterized and constrained within predefined bounds. This configuration, including both the network architecture and parameter setting, is based on that proposed in \cite{10251604}. 

The overall loss function combines a data-driven loss term and a physics-based loss term, weighted by $\lambda_u$ and $\lambda_f$, both of them set to 1 after analyzing their value scales.

In addition, a conventional FNN is trained as a baseline to evaluate the impact of incorporating physical constraints. This comparison allows the advantages of the proposed PINN-based pipeline to be quantitatively assessed.

Table \ref{tab:results} reports the loss values obtained for each discharge segment on the validation battery set. 

\begin{table}[H]
\caption{Error metrics for SOH prediction across discharge segments.}
\label{tab:results}
\centering
\begin{tabular}{ccccccc}
\toprule
& \multicolumn{2}{c}{\textbf{MAPE (\%)}} & \multicolumn{2}{c}{\textbf{RMSE}} & \multicolumn{2}{c}{\textbf{R$^2$}} \\
\cmidrule(lr){2-3}\cmidrule(lr){4-5}\cmidrule(lr){6-7}
\textbf{Segment} & \textbf{PINN} & \textbf{FNN} & \textbf{PINN} & \textbf{FNN}& \textbf{PINN} & \textbf{FNN} \\
\midrule
1 & 3.096 & 4.239 & 0.038 & 0.049 & 0.734 & 0.567 \\
2 & 3.835 & 5.046 & 0.047 & 0.057 & 0.594 & 0.401 \\
3 & 3.714 & 4.338 & 0.046 & 0.050 & 0.609 & 0.553 \\
4 & 2.867 & 4.678 & 0.036 & 0.053 & 0.766 & 0.492 \\
5 & 3.405 & 5.992 & 0.040 & 0.063 & 0.709 & 0.275 \\
6 & 1.543 & 2.229 & 0.016 & 0.023 & 0.952 & 0.902 \\
7 & 3.094 & 4.649 & 0.034 & 0.049 & 0.791 & 0.557 \\
8 & 2.245 & 3.098 & 0.025 & 0.034 & 0.887 & 0.795 \\
\bottomrule
\end{tabular}
\end{table}

An analysis of Table \ref{tab:results} shows that, for every discharge segment and across all evaluation metrics, the PINN consistently outperforms the conventional NN, yielding more accurate SOH predictions. Overall, none of the segments exhibits a MAPE greater than 4$\%$, indicating good predictive performance across the entire voltage range.

Among all segments, the sixth segment achieves the highest prediction accuracy, with a MAPE of 1.52$\%$, an RMSE of 0.016, and an $R^2$ of 0.952. This performance can be attributed to the fact that the sixth segment covers voltage values around 3.14 V, which are close to the bend point of the discharge curve for this nominal voltage range. Although the model is not explicitly provided with the location of the bend point, the information contained in this voltage region is highly informative for SOH estimation, leading to improved predictive accuracy. In addition, as shown in Table \ref{tab:feature_selection}, this segment extracts the relevant information using only the slope of the VQ curve, enabling a lightweight model that relies on a single predictor and avoids the computation of additional features.

Likewise, the eighth segment, corresponding to the beginning of the discharge curve, also exhibits good performance, with a MAPE of 2.21$\%$, an RMSE of 0.025 and an $R^2$ of 0.887.

In contrast, the poorest performance is observed for the second and third segments, with MAPEs of 3.835$\%$ and 3.714$\%$, RMSE values of 0.047 and 0.046, and $R^2$ values of 0.594 and 0.609, respectively. Nevertheless, both segments still achieve MAPEs below 4$\%$, indicating acceptable predictive accuracy. Table \ref{tab:results} addresses \textbf{RQ2}, demonstrating that the proposed framework can accurately estimate the SOH from arbitrary discharge segments without requiring complete discharge curves or predefined voltage regions. The XGBoost classifier automatically assigns an unlabeled discharge segment to its corresponding voltage-range category, allowing the appropriate PINN to be selected and enabling the framework to operate without prior knowledge of the segment location within the discharge curve.

\begin{figure}[h!]
    \centering
    \includegraphics[width=0.9\columnwidth]{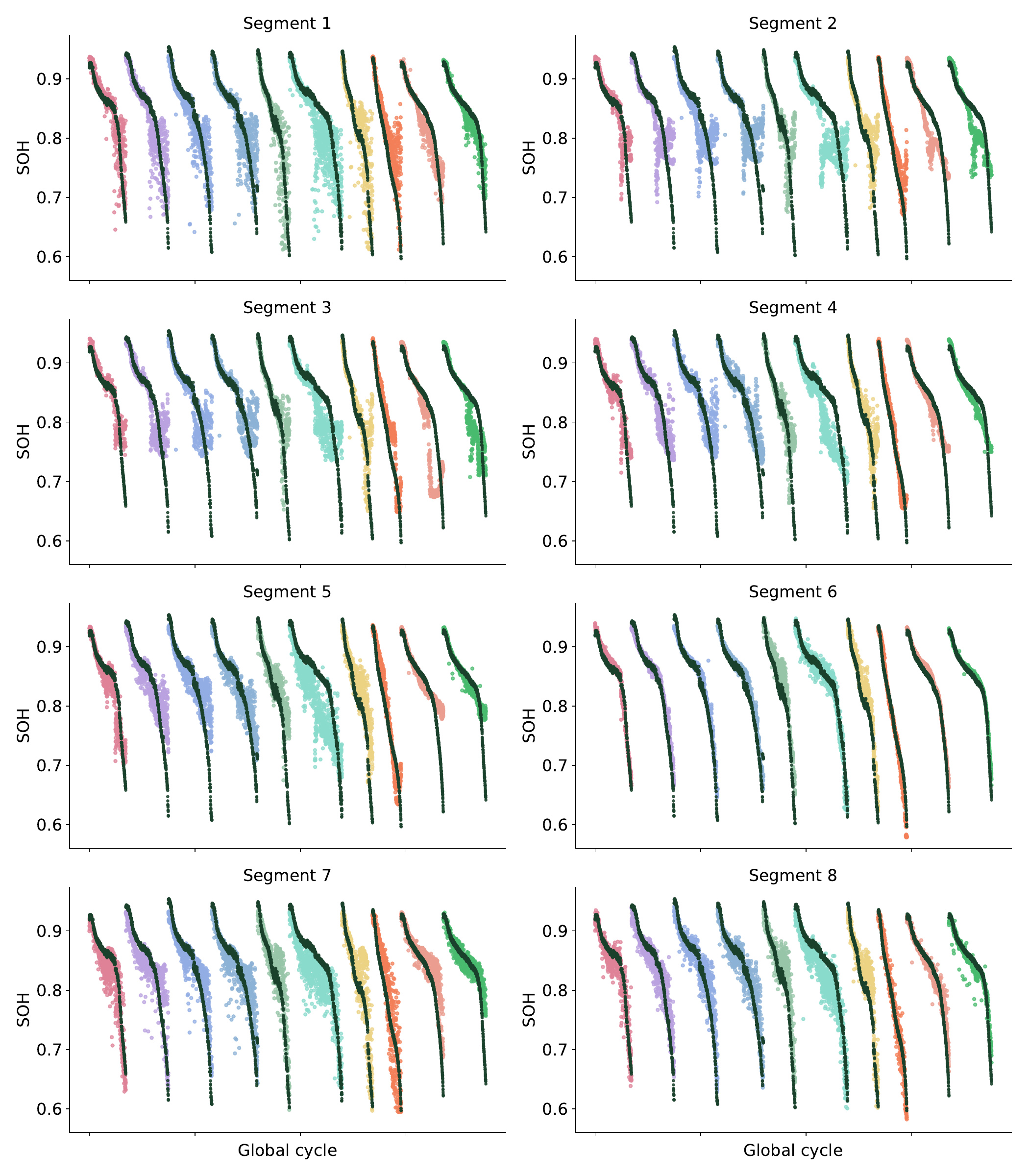}
    \caption{Predicted vs. ground-truth SOH for each discharge segment in the validation set}
    \label{fig:validación}
\end{figure}

Figure \ref{fig:validación} illustrates the capacity degradation of the batteries reserved for validation across different segments, together with the corresponding predictions obtained by each model. 

Dark points indicate the ground-truth SOH values, while each colored marker corresponds to an individual battery in the validation set. The same color across different segment plots represents the same battery.The horizontal axis denotes the accumulated cycle count, concatenated sequentially across batteries, and the vertical axis represents the SOH. 

This representation highlights the diversity of degradation trajectories across batteries, including differences in shape and lifespan. It also demonstrates that the proposed models are capable of capturing changes in the degradation trend. Although the prediction of the second and third segments exhibit the largest errors, these models are still able to distinguish and follow the different degradation behaviors among batteries. Consistently with the quantitative results reported in Table \ref{tab:results}, the sixth and eighth segments yield the best overall performance.

After obtaining the SOH predictions, a real-time estimation of the battery degradation trend is also proposed. This approach provides the user with an indicator that reflects how the battery is expected to degrade, together with an associated confidence interval. The degradation trend is modeled using linear regressions that are continuously updated as new SOH estimates become available. The CI is computed from the variability of the slopes estimated over the last 100 updated regression models. The window length, $\phi=100$, was determined experimentally, as this corresponds to the point at which the standard deviation of the slope estimates exhibits a noticeable change in trend.

When new predictions lie outside the CI, the knee-point of the degradation curve, and consequently a change in the degradation regime, are expected to be identified. To determine the cycle at which the knee-point occurs, analyzed the number of consecutive SOH predictions that fall outside the confidence interval of the estimated degradation trend is analyzed.

\begin{figure}[h!]
    \centering
    \includegraphics[width=0.4\columnwidth]{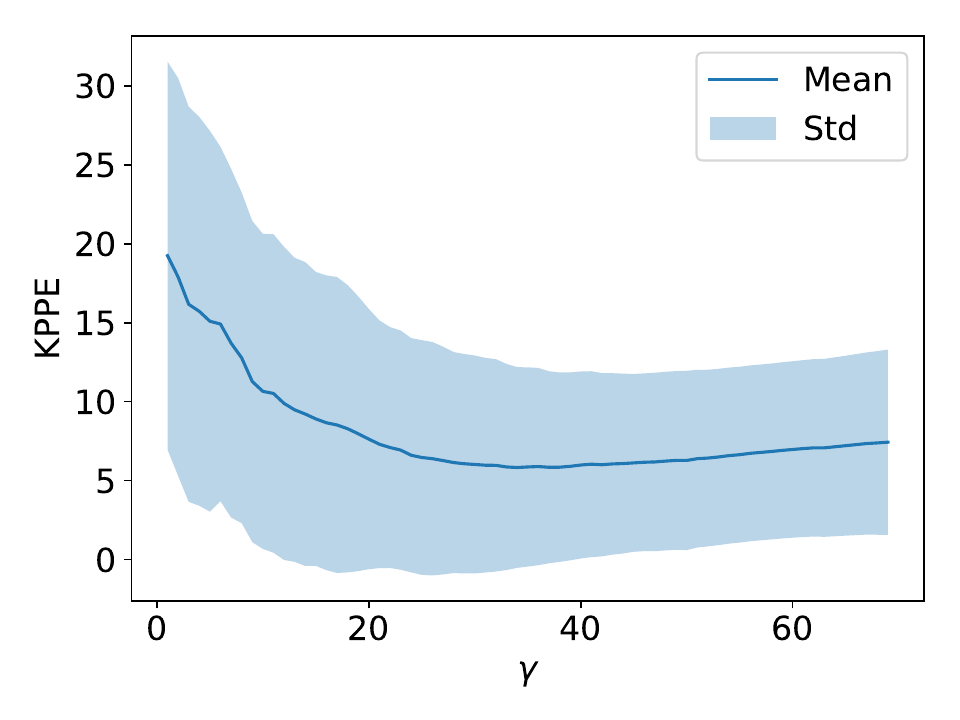}
    \caption{KPPE value for each $\gamma$}
    \label{fig:kppe}
\end{figure}

Figure \ref{fig:kppe} reports the distance between the estimated knee-point and the ground-truth knee-point, as defined in Equation \ref{eq:KPPE}. In particular, it shows to the mean and standard deviation of the KPPE computed across all batteries for each $\gamma$.

$\gamma$ with the lowest KPPE value can be also inferred from the Figure \ref{fig:kppe}. For the considered dataset, the minimum mean error is achieved when 34 consecutive SOH predictions fall outside the CI. Therefore, $\gamma = 34$ is selected to declare the knee point, indicating the transition to an accelerated degradation regime.

\begin{figure}[h!]
    \centering
    \includegraphics[width=1\columnwidth]{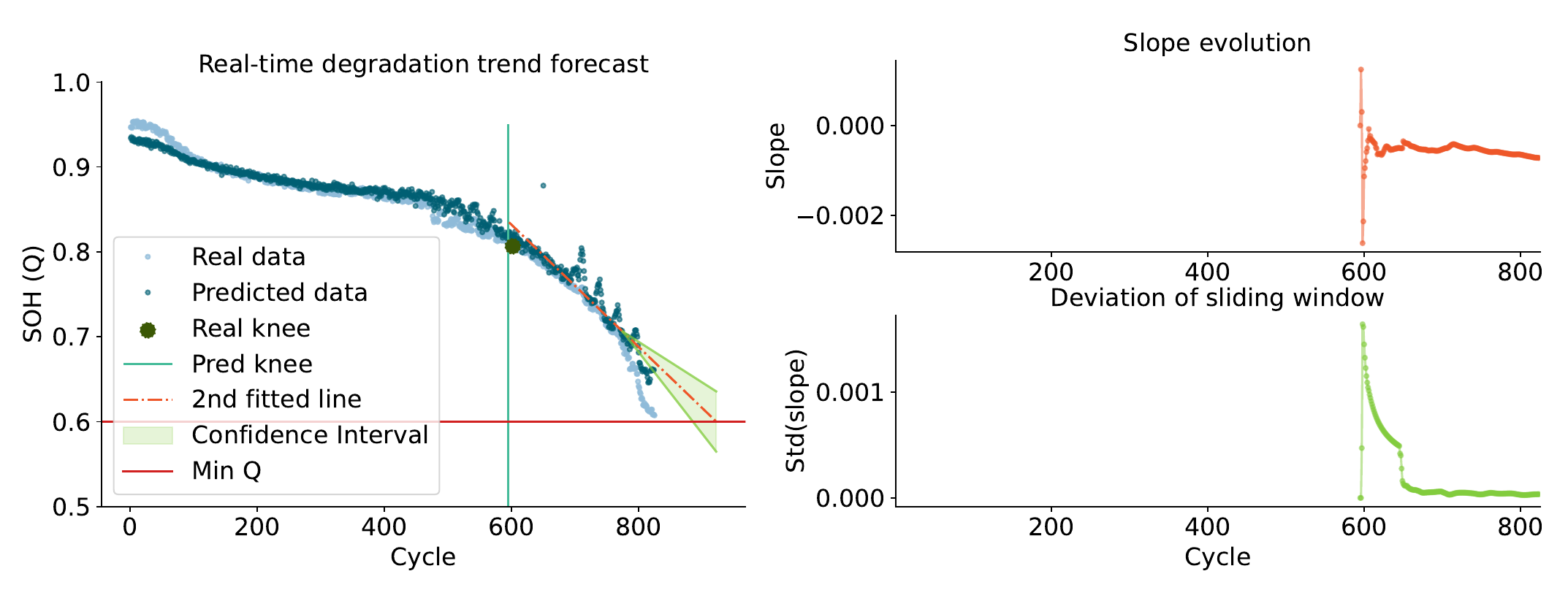}
    \caption{Degradation trend estimation and stability analysis. Left: fitted degradation line (dashed orange) with its corresponding confidence interval (shaded green) and predicted SOH values (green). Right: temporal evolution of the slope (top) and the standard deviation of the slope computed over a sliding window of 50 cycles (bottom).}
    \label{fig:second_regime_approach}
\end{figure}

From this point onward, subsequent SOH predictions are used to characterize the accelerated degradation regime. This second degradation trend is fitted using a least-squares regression and extrapolated until a predefined minimum capacity threshold, specified by the user, is reached. This is represented in Figure \ref{fig:second_regime_approach}.

As in the previous regime, a CI is defined based on the slopes of the last 50 time steps, where the change in slope becomes evident, as shown in the bottom-right plot. In addition, during the accelerated degradation phase, the predictions are not as accurate as in the first regime. This is due to the different lifespans among batteries, which makes it difficult for the network to learn SOH values after a certain number of cycles. The cycles at which the confidence interval intersects the threshold are considered possible EoL points for the battery, triggering the second alarm. However, as mentioned before, the minimum capacity value is up to the user, thus, this second approach would be generalized for any other minimum value. 

With these final developments, \textbf{RQ4} can now be addressed. This work has successfully designed, first, a PINN capable of estimating the battery SOH using only a voltage‑based segment taken at a random point of the discharge curve. The analysis has demonstrated the central role of the discharge process compared with the charge process. In addition, features were extracted for each voltage segment, showing that different voltage ranges encode predictive information in different variables.
Furthermore, building on the SOH approximation, a real‑time model has been developed to estimate the battery degradation trajectory, including an alarm mechanism that detects the knee‑point, reports the change in degradation trend, and provides an approximation of the battery EoL.

The real-time performance of the proposed approach is demonstrated in a supplementary video available.

\section{Conclusions}
\label{conclusions}

The widespread adoption of lithium-ion batteries in recent years has intensified the demand for enhanced energy storage performance. In this context, the present study aims to explore the prediction of the SOH of these batteries based on their physical degradation dynamics, as well as the estimation of SOH decay trends. The proposed methodology has been trained using a dataset comprising 122 lithium-ion batteries, of which 10 were exclusively reserved for testing purposes.

This work proposes a method that, addressing \textbf{RQ2}, generalizes across different voltage ranges of the discharge curve and, in line with \textbf{RQ4}, leverages the resulting SOH estimate to characterize the degradation trend of the battery. Specifically, the study introduces a framework capable of estimating the SOH from an arbitrary discharge segment and subsequently using this estimate for real‑time degradation prediction. To this end, multiple deep learning models were developed, each corresponding to one of eight uniformly distributed voltage ranges between 2.0 V and 3.6 V, which span the operational range of the available batteries. These models were trained independently, resulting in the use of distinct predictors for each voltage section. Starting from an initial feature set, predictors relevance was evaluated using Random Forest–based feature importance for each section, and the most informative predictors were selected for each segment. The analysis addresses \textbf{RQ3}, demonstrating that different voltage range has relevant information hidden on distinct features. 

Based on these selected predictors, physics-informed models were developed by incorporating the Verhulst equation, which has been shown to accurately represent the degradation process of lithium-ion batteries, along with a loss term based on the error between the predicted SOH and the ground truth. This formulation ensures the convergence towards the ground truth while maintaining the physical consistency.

Among the developed models, the one corresponding to the voltage segment centered around 3.14 V achieved the highest accuracy. This result is consistent with the fact that this voltage range is located near the knee of the discharge curve, a region that is highly informative of the battery SOH, and it depends only on the slope of the discharge curve. 
Hence, this result ensures the robustness of the method validating the generalization capability of the method for predicting the SOH of the battery when arbitrary discharge segments are available (\textbf{RQ2}).

These predictions are then utilized to estimate real-time degradation trend, as addressed in \textbf{RQ4}. The predicted SOH values are fitted using a least-squares linear regression, from which a CI is constructed. This CI is then used to identify the knee-point of the degradation, determining the beginning of a second degradation regime, which is estimated to end when the corresponding CI intersects the minimum valid capacity value. 

Overall, the proposed approach enables an accurate, battery‑agnostic estimation of the degradation trend, applicable to individual batteries without requiring any prior information. Moreover, unlike traditional early‑based prediction methods, this real‑time estimator incorporates every predicted SOH value and thus captures the evolving behavior of each battery, yielding a more reliable and adaptive characterization of its degradation trajectory.

\section*{Future work}\label{future_work}
Several directions for future work emerge from this study. First, the optimal number of models\textemdash that is, the number of segments\textemdash required to effectively cover the entire discharge curve should be investigated, together with the optimal length of each segment. The generalization of the proposed pipeline to more heterogeneous battery datasets should also be evaluated. In addition, the performance of the proposed approach on batteries operating under different charging and discharging current profiles should be assessed. The optimization of the Verhulst model parameters in scenarios involving multiple battery types also deserves further investigation. Furthermore, the integration of additional physical variables, such as temperature and pressure, into the Verhulst model could be explored to further improve SOH prediction. Finally, future work should investigate strategies for combining multiple NN models when the available discharge segment spans a wider voltage range than expected, including the development of appropriate model-weighting schemes.

\bibliographystyle{cas-model2-names}
\bibliography{sample}

@article{SORDI2025100410,
title = {Degradation of lithium-ion batteries under automotive-like conditions: P2D model-based understanding and ex-situ validation},
journal = {eTransportation},
volume = {24},
pages = {100410},
year = {2025},
issn = {2590-1168},
doi = {https://doi.org/10.1016/j.etran.2025.100410},
url = {https://www.sciencedirect.com/science/article/pii/S2590116825000177},
author = {G. Sordi and A. Stecchini and R. Evangelista and D. Luder and W. Li and D.U. Sauer and A. Casalegno and C. Rabissi}

}

@article{Karger2024,
  title   = {Modeling Particle Versus SEI Cracking in Lithium-Ion Battery Degradation: Why Calendar and Cycle Aging Cannot Simply be Added},
  author  = {Karger, Alexander and O’Kane, Simon E. J. and Rogge, Marcel and Kirst, Cedric and Singer, Jan P. and Marinescu, Monica and Offer, Gregory J. and Jossen, Andreas},
  journal = {Journal of The Electrochemical Society},
  volume  = {171},
  number  = {9},
  pages   = {090512},
  year    = {2024},
  doi     = {10.1149/1945-7111/ad76da},
  url     = {https://doi.org/10.1149/1945-7111/ad76da}
}

@Article{batteries6030037,
AUTHOR = {Al-Gabalawy, Mostafa and Hosny, Nesreen S. and Hussien, Shimaa A.},
TITLE = {Lithium-Ion Battery Modeling Including Degradation Based on Single-Particle Approximations},
JOURNAL = {Batteries},
VOLUME = {6},
YEAR = {2020},
NUMBER = {3},
ARTICLE-NUMBER = {37},
URL = {https://www.mdpi.com/2313-0105/6/3/37},
ISSN = {2313-0105},
DOI = {10.3390/batteries6030037}
}

@article{FERMINCUETO2020100006,
title = {Identification and machine learning prediction of knee-point and knee-onset in capacity degradation curves of lithium-ion cells},
journal = {Energy and AI},
volume = {1},
pages = {100006},
year = {2020},
issn = {2666-5468},
doi = {https://doi.org/10.1016/j.egyai.2020.100006},
url = {https://www.sciencedirect.com/science/article/pii/S2666546820300069},
author = {Fermín-Cueto, Paula and McTurk, Euan and Allerhand, Michael and Medina-Lopez, Encarni and F. Anjos, Miguel and Sylvester, Joel and dos Reis,  Gonçalo}
}

@ARTICLE{9520291,
  author={Greenbank, Samuel and Howey, David},
  journal={IEEE Transactions on Industrial Informatics}, 
  title={Automated Feature Extraction and Selection for Data-Driven Models of Rapid Battery Capacity Fade and End of Life}, 
  year={2022},
  volume={18},
  number={5},
  pages={2965-2973},
  doi={10.1109/TII.2021.3106593}
}

@article{Attia2022Knees,
  title   = {Review---“Knees” in Lithium-Ion Battery Aging Trajectories},
  author  = {Attia, Peter M. and Bills, Alexander and Brosa Planella, Ferran and Dechent, Philipp and dos Reis, Gonçalo and Dubarry, Matthieu and Gasper, Paul and Gilchrist, Richard and Greenbank, Samuel and Howey, David},
  journal = {Journal of The Electrochemical Society},
  volume  = {169},
  number  = {6},
  pages   = {060517},
  year    = {2022},
  publisher = {IOP Publishing Ltd on behalf of The Electrochemical Society},
  doi     = {10.1149/1945-7111/ac6d13}
}

@article{ECM_p,
title = "Constrained Ensemble Kalman Filter for Distributed Electrochemical State Estimation of Lithium-Ion Batteries",
author = "Yang Li and Binyu Xiong and Vilathgamuwa, \{Don Mahinda\} and Zhongbao Wei and Changjun Xie and Changfu Zou",
note = "Publisher Copyright: {\textcopyright} 2005-2012 IEEE.",
year = "2021",
month = jan,
doi = "10.1109/TII.2020.2974907",
language = "English",
volume = "17",
pages = "240--250",
journal = "IEEE Transactions on Industrial Informatics",
issn = "1551-3203",
publisher = "IEEE Computer Society",
number = "1",
}

@Article{batteries11050167,
AUTHOR = {He, Shengfeng and Qin, Wenhu and Yun, Zhonghua and Wu, Chao and Sun, Chongbin},
TITLE = {SOH Estimation Method for Lithium-Ion Batteries Using Partial Discharge Curves Based on CGKAN},
JOURNAL = {Batteries},
VOLUME = {11},
YEAR = {2025},
NUMBER = {5},
ARTICLE-NUMBER = {167},
URL = {https://www.mdpi.com/2313-0105/11/5/167},
ISSN = {2313-0105},

DOI = {10.3390/batteries11050167}
}

@article{LI2019510,
title = {Remaining useful life prediction for lithium-ion batteries based on a hybrid model combining the long short-term memory and Elman neural networks},
journal = {Journal of Energy Storage},
volume = {21},
pages = {510-518},
year = {2019},
issn = {2352-152X},
doi = {https://doi.org/10.1016/j.est.2018.12.011},
url = {https://www.sciencedirect.com/science/article/pii/S2352152X1830450X},
author = {Xiaoyu Li and Lei Zhang and Zhenpo Wang and Peng Dong}
}

@article{Fetanat2023FENN,
  title   = {Fully Elman Neural Network: A Novel Deep Recurrent Neural Network Optimized by an Improved Harris Hawks Algorithm for Classification of Pulmonary Arterial Wedge Pressure},
  author  = {Fetanat, Masoud and Stevens, Michael and Jain, Pankaj and Hayward, Christopher and Meijering, Erik and Lovell, Nigel H.},
  journal = {arXiv preprint arXiv:2301.07710},
  year    = {2023},
  doi     = {10.48550/arXiv.2301.07710},
  archivePrefix = {arXiv},
  eprint  = {2301.07710},
  primaryClass = {cs.LG}
}

@article{LI20181178,
title = {A single particle model with chemical/mechanical degradation physics for lithium ion battery State of Health (SOH) estimation},
journal = {Applied Energy},
volume = {212},
pages = {1178-1190},
year = {2018},
issn = {0306-2619},
doi = {https://doi.org/10.1016/j.apenergy.2018.01.011},
url = {https://www.sciencedirect.com/science/article/pii/S0306261918300114},
author = {J. Li and K. Adewuyi and N. Lotfi and R.G. Landers and J. Park}
}

@ARTICLE{9137406,
  author={Ren, Lei and Dong, Jiabao and Wang, Xiaokang and Meng, Zihao and Zhao, Li and Deen, M. Jamal},
  journal={IEEE Transactions on Industrial Informatics}, 
  title={A Data-Driven Auto-CNN-LSTM Prediction Model for Lithium-Ion Battery Remaining Useful Life}, 
  year={2021},
  volume={17},
  number={5},
  pages={3478-3487},
  doi={10.1109/TII.2020.3008223}}

@article{Xu2024PhysicsAwareML,
  title        = {Physics-aware Machine Learning Revolutionizes Scientific Paradigm for Machine Learning and Process-based Hydrology},
  author       = {Xu, Qingsong and Shi, Yilei and Bamber, Jonathan and Tuo, Ye and Ludwig, Ralf and Zhu, Xiao Xiang},
  journal      = {arXiv preprint arXiv:2310.05227},
  year         = {2024},
  note         = {Submitted on 8 Oct 2023, last revised 12 Jul 2024 (v5)},
  url          = {https://doi.org/10.48550/arXiv.2310.05227},
  doi          = {10.48550/arXiv.2310.05227}
}

@article{RAISSI2019686,
title = {Physics-informed neural networks: A deep learning framework for solving forward and inverse problems involving nonlinear partial differential equations},
journal = {Journal of Computational Physics},
volume = {378},
pages = {686-707},
year = {2019},
issn = {0021-9991},
doi = {https://doi.org/10.1016/j.jcp.2018.10.045},
url = {https://www.sciencedirect.com/science/article/pii/S0021999118307125},
author = {M. Raissi and P. Perdikaris and G.E. Karniadakis}
}

@article{Long2022AutoIP,
  title   = {AutoIP: A Unified Framework to Integrate Physics into Gaussian Processes},
  author  = {Long, Da and Wang, Zheng and Krishnapriyan, Aditi and Kirby, Robert and Zhe, Shandian and Mahoney, Michael},
  journal = {arXiv preprint arXiv:2202.12316},
  year    = {2022},
  doi     = {10.48550/arXiv.2202.12316},
  archivePrefix = {arXiv},
  eprint  = {2202.12316},
  primaryClass = {cs.LG}
}

@article{Wang2021PIDeepONet,
  title   = {Learning the Solution Operator of Parametric Partial Differential Equations with Physics-Informed DeepONets},
  author  = {Wang, Sifan and Wang, Hanwen and Perdikaris, Paris},
  journal = {arXiv preprint arXiv:2103.10974},
  year    = {2021},
  doi     = {10.48550/arXiv.2103.10974},
  archivePrefix = {arXiv},
  eprint  = {2103.10974},
  primaryClass = {cs.LG}
}

@article{Wang2024PINN,
  title   = {Physics-informed neural network for lithium-ion battery degradation stable modeling and prognosis},
  author  = {Wang, Feng and Zhai, Zhiyuan and Zhao, Zhen and others},
  journal = {Nature Communications},
  volume  = {15},
  number  = {},
  pages   = {4332},
  year    = {2024},
  issn    = {2041-1723},
  doi     = {10.1038/s41467-024-48779-z},
  url     = {https://doi.org/10.1038/s41467-024-48779-z}
}

@article{TANG2025825,
title = {Physics-informed battery degradation prediction: Forecasting charging curves using one-cycle data},
journal = {Journal of Energy Chemistry},
volume = {101},
pages = {825-836},
year = {2025},
issn = {2095-4956},
doi = {https://doi.org/10.1016/j.jechem.2024.10.018},
url = {https://www.sciencedirect.com/science/article/pii/S2095495624007204},
author = {Aihua Tang and Yuchen Xu and Jinpeng Tian and Xing Shu and Quanqing Yu}
}

@Inbook{Bacaer2011,
author="Baca{\"e}r, Nicolas",
title="Verhulst and the logistic equation (1838)",
bookTitle="A Short History of Mathematical Population Dynamics",
year="2011",
publisher="Springer London",
address="London",
pages={35--39},
isbn="978-0-85729-115-8",
doi="10.1007/978-0-85729-115-8_6",
url="https://doi.org/10.1007/978-0-85729-115-8_6"
}

@ARTICLE{10251604,
  author={Wen, Pengfei and Ye, Zhi-Sheng and Li, Yong and Chen, Shaowei and Xie, Pu and Zhao, Shuai},
  journal={IEEE Transactions on Intelligent Vehicles}, 
  title={Physics-Informed Neural Networks for Prognostics and Health Management of Lithium-Ion Batteries}, 
  year={2024},
  volume={9},
  number={1},
  pages={2276-2289},
  doi={10.1109/TIV.2023.3315548}}

@Inbook{Maafi2025,
author="Maafi, Mounir",
title="Established Semi-Emperical Model Equations (SEM)",
bookTitle="Photokinetics: A New Perspective",
year="2025",
publisher="Springer Nature Switzerland",
address={Cham},
pages={277--292},
isbn="978-3-031-98985-8",
doi="10.1007/978-3-031-98985-8_15",
url="https://doi.org/10.1007/978-3-031-98985-8_15"
}

@article{SUN2025126425,
title = {A degradation trajectory prediction method applicable to various life stages of lithium-ion batteries under complex variable aging conditions},
journal = {Applied Energy},
volume = {398},
pages = {126425},
year = {2025},
issn = {0306-2619},
doi = {https://doi.org/10.1016/j.apenergy.2025.126425},
url = {https://www.sciencedirect.com/science/article/pii/S0306261925011559},
author = {Jinghua Sun and Jiajie Lou and Josef Kainz}
}

@ARTICLE{10927632,
  author={Wang, Lingchen and Yang, Tao and Hu, Bo},
  journal={IEEE Sensors Journal}, 
  title={A Battery State-of-Health Estimation Method for Real-World Electric Vehicles Based on Physics-Informed Neural Networks}, 
  year={2025},
  volume={25},
  number={9},
  pages={15577-15587},
  doi={10.1109/JSEN.2025.3549486}}

@ARTICLE{9454160,
  author={Hasib, Shahid A. and Islam, S. and Chakrabortty, Ripon K. and Ryan, Michael J. and Saha, D. K. and Ahamed, Md H. and Moyeen, S. I. and Das, Sajal K. and Ali, Md F. and Islam, Md R. and Tasneem, Z. and Badal, Faisal R.},
  journal={IEEE Access}, 
  title={A Comprehensive Review of Available Battery Datasets, RUL Prediction Approaches, and Advanced Battery Management}, 
  year={2021},
  volume={9},
  number={},
  pages={86166-86193},
  doi={10.1109/ACCESS.2021.3089032}}

@Article{en13092380,
AUTHOR = {Mao, Ling and Xu, Jie and Chen, Jiajun and Zhao, Jinbin and Wu, Yuebao and Yao, Fengjun},
TITLE = {A LSTM-STW and GS-LM Fusion Method for Lithium-Ion Battery RUL Prediction Based on EEMD},
JOURNAL = {Energies},
VOLUME = {13},
YEAR = {2020},
NUMBER = {9},
ARTICLE-NUMBER = {2380},
URL = {https://www.mdpi.com/1996-1073/13/9/2380},
ISSN = {1996-1073},
DOI = {10.3390/en13092380}
}

@article{SUN2021108679,
title = {A hybrid prognostic strategy with unscented particle filter and optimized multiple kernel relevance vector machine for lithium-ion battery},
journal = {Measurement},
volume = {170},
pages = {108679},
year = {2021},
issn = {0263-2241},
doi = {https://doi.org/10.1016/j.measurement.2020.108679},
url = {https://www.sciencedirect.com/science/article/pii/S0263224120311908},
author = {Xiaofei Sun and Kai Zhong and Min Han}
}

@article{CHEN2024110471,
title = {A hybrid battery degradation model combining arrhenius equation and neural network for capacity prediction under time-varying operating conditions},
journal = {Reliability Engineering and System Safety},
volume = {252},
pages = {110471},
year = {2024},
issn = {0951-8320},
doi = {https://doi.org/10.1016/j.ress.2024.110471},
url = {https://www.sciencedirect.com/science/article/pii/S095183202400543X},
author = {Zhen Chen and Zirong Wang and Wei Wu and Tangbin Xia and Ershun Pan}
}

@ARTICLE{6587560,
  author={Xian, Weiming and Long, Bing and Li, Min and Wang, Houjun},
  journal={IEEE Transactions on Instrumentation and Measurement}, 
  title={Prognostics of Lithium-Ion Batteries Based on the Verhulst Model, Particle Swarm Optimization and Particle Filter}, 
  year={2014},
  volume={63},
  number={1},
  pages={2-17},
  doi={10.1109/TIM.2013.2276473}}

@article{Xu2021,
  title   = {Comparison of the effect of linear and two-step fast charging protocols on degradation of lithium-ion batteries},
  author  = {Xu, M. and Wang, X. and Zhang, L. and Zhao, P.},
  journal = {Energy},
  volume  = {227},
  pages   = {120417},
  year    = {2021},
  issn    = {0360-5442},
  doi     = {10.1016/j.energy.2021.120417}
}

@article{PETRELLA2025118747,
title = {Fast estimation of lithium-ion battery state of health using time series classification},
journal = {Journal of Energy Storage},
volume = {138},
pages = {118747},
year = {2025},
issn = {2352-152X},
doi = {https://doi.org/10.1016/j.est.2025.118747},
url = {https://www.sciencedirect.com/science/article/pii/S2352152X25034607},
author = {Alessandro Petrella and Moreno Marzolla and Francesco Mercuri}
}

@article{YANG2022103857,
title = {Robust State of Health estimation of lithium-ion batteries using convolutional neural network and random forest},
journal = {Journal of Energy Storage},
volume = {48},
pages = {103857},
year = {2022},
issn = {2352-152X},
doi = {https://doi.org/10.1016/j.est.2021.103857},
url = {https://www.sciencedirect.com/science/article/pii/S2352152X21015231},
author = {Niankai Yang and Ziyou Song and Heath Hofmann and Jing Sun}
}

@ARTICLE{8603757,
  author={Anseán, David and García, Víctor Manuel and González, Manuela and Blanco-Viejo, Cecilio and Viera, Juan Carlos and Pulido, Yoana Fernández and Sánchez, Luciano},
  journal={IEEE Transactions on Industry Applications}, 
  title={Lithium-Ion Battery Degradation Indicators Via Incremental Capacity Analysis}, 
  year={2019},
  volume={55},
  number={3},
  pages={2992-3002},
  doi={10.1109/TIA.2019.2891213}}

@InProceedings{RF_featu_importance,
author="Tang, Adelina
and Foong, Joan Tack",
editor="Herawan, Tutut and Ghazali, Rozaida and Deris, Mustafa Mat",
title="A Qualitative Evaluation of Random Forest Feature Learning",
booktitle="Recent Advances on Soft Computing and Data Mining",
year="2014",
publisher="Springer International Publishing",
address="Cham",
pages="359--368",
isbn="978-3-319-07692-8"
}

@article{S2024123542,
title = {State of Health (SoH) estimation methods for second life lithium-ion battery—Review and challenges},
journal = {Applied Energy},
volume = {369},
pages = {123542},
year = {2024},
issn = {0306-2619},
doi = {https://doi.org/10.1016/j.apenergy.2024.123542},
url = {https://www.sciencedirect.com/science/article/pii/S0306261924009255},
author = { Vignesh, S and Hang Seng Che and Jeyraj Selvaraj and Kok Soon Tey and Jia Woon Lee and Hussain Shareef and Rachid Errouissi}
}

@article{Severson2019,
  title   = {Data-driven prediction of battery cycle life before capacity degradation},
  author  = {Severson, Kristen A. and Attia, Peter M. and Jin, Ning and Perkins, Nicholas and Jiang, Billy and Yang, Zi and Chen, Matthew H. and Aykol, Muratahan and Herring, Patrick K. and Fraggedakis, Dimitrios and others},
  journal = {Nature Energy},
  volume  = {4},
  number  = {5},
  pages   = {383--391},
  year    = {2019},
  doi     = {10.1038/s41560-019-0356-8},
  url     = {https://doi.org/10.1038/s41560-019-0356-8}
}

@article{Breiman2001,
  author    = {Breiman, Leo},
  title     = {Random Forests},
  journal   = {Machine Learning},
  volume    = {45},
  number    = {1},
  pages     = {5--32},
  year      = {2001},
  doi       = {10.1023/A:1010933404324},
  url       = {https://doi.org/10.1023/A:1010933404324},
  issn      = {1573-0565}
}

@inproceedings{Chen2016XGBoost,
  author    = {Chen, Tianqi and Guestrin, Carlos},
  title     = {XGBoost: A Scalable Tree Boosting System},
  booktitle = {Proceedings of the 22nd ACM SIGKDD International Conference on Knowledge Discovery and Data Mining},
  pages     = {785--794},
  year      = {2016},
  publisher = {ACM},
  doi       = {10.1145/2939672.2939785},
  url       = {https://doi.org/10.1145/2939672.2939785}
}

@article{Rumelhart1986Backprop,
  author  = {Rumelhart, David E. and Hinton, Geoffrey E. and Williams, Ronald J.},
  title   = {Learning representations by back-propagating errors},
  journal = {Nature},
  volume  = {323},
  number  = {6088},
  pages   = {533--536},
  year    = {1986},
  doi     = {10.1038/323533a0},
  url     = {https://doi.org/10.1038/323533a0}
}

@INPROCEEDINGS{begoIbai,
  author={Ispizua, Begoña and Gil-Lopez, Sergio and Laña, Ibai},
  booktitle={2025 10th International Conference on Power and Renewable Energy (ICPRE)}, 
  title={The Role of Discharge Curve Knee Point Analysis for Early Li-ion Battery End of Life Forecasting}, 
  year={2025},
  volume={},
  number={},
  pages={1006-1011},
  doi={10.1109/ICPRE67300.2025.11274154}}

@INPROCEEDINGS{5961514,
  author={Satopaa, Ville and Albrecht, Jeannie and Irwin, David and Raghavan, Barath},
  booktitle={2011 31st International Conference on Distributed Computing Systems Workshops}, 
  title={Finding a "Kneedle" in a Haystack: Detecting Knee Points in System Behavior}, 
  year={2011},
  volume={},
  number={},
  pages={166-171},
  doi={10.1109/ICDCSW.2011.20}}

\section*{Acknowledgments}
This work was supported by the BAT4ME project (ref. KK-2023/00048) funded by the Department of Industry of the Basque Government through the Collaborative Research Grants Programme in strategic areas - Elkartek Programme of the 2023 call.

This work was also supported by a grant from the Department of Science, University and Innovation from the Basque Government to the MATHMODE Group (IT1866-26).

\section*{Declaration of competing interest}
The authors declare that they have no known competing financial interests or personal relationships that could have appeared to influence the work reported in this paper.

\section*{CRediT author statement}

\textbf{CRediT}: Begoña Ispizua: Conceptualization, Investigation, Methodology, Writing – original draft, review \& editing; Sergio Gil-López: Conceptualization, Methodology, Supervision; Leire Arrizabalaga: Software; Ibai Laña: Conceptualization, Supervision, Writing – review \& editing.

\vskip3pt

\begin{figure}[h!]
    \centering
    \includegraphics[width=0.2\columnwidth]{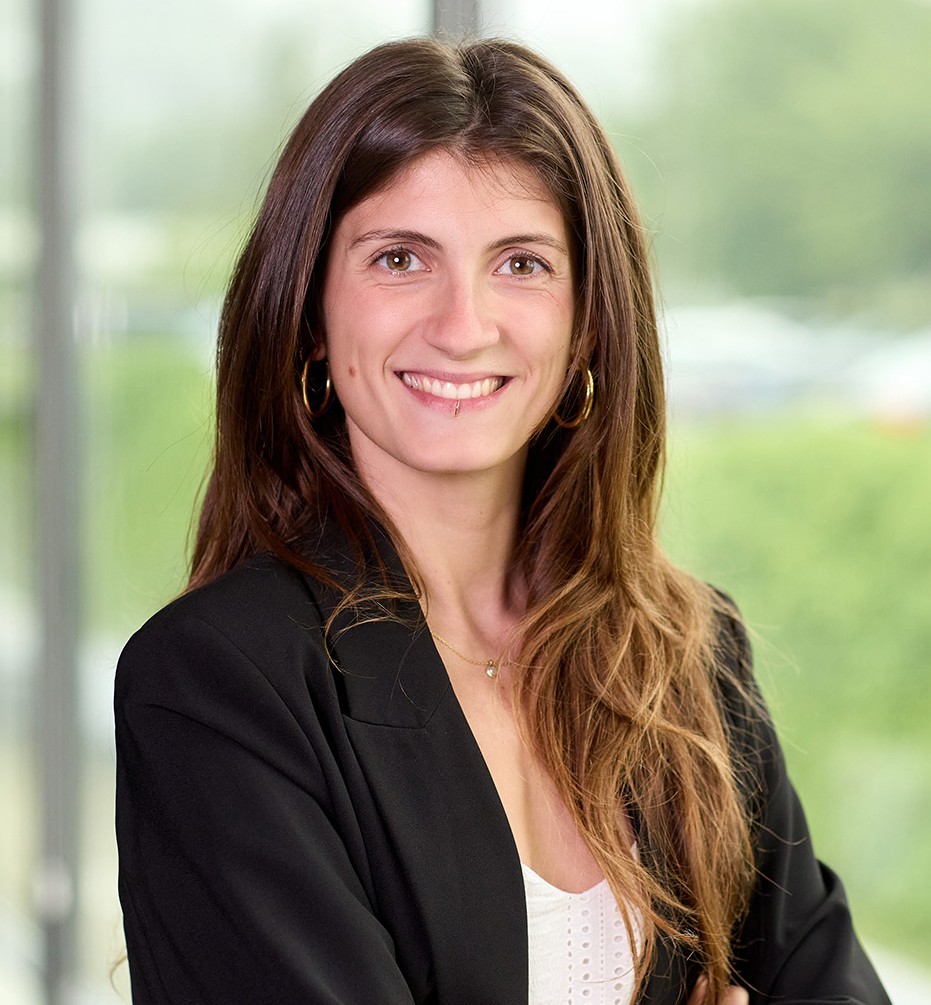}
\end{figure}
\textbf{Begoña Ispizua} received her B.Sc. degree in Mathematics from the University of the Basque Country (EHU), where she also obtained her M.Sc. in Mathematical Modeling, Statistics, and Computational Research. She has been pursuing her Ph.D. since 2024 while working as a researcher at Tecnalia. Her work focuses on the application of artificial intelligence across several domains, including healthcare, industry, and energy. In particular, she has experience in reinforcement learning (RL), anomaly detection, and physics-informed machine learning. Currently, her research is centered on the estimation of the remaining useful life of batteries.

\vskip2pc

\begin{figure}[h!]
    \centering
    \includegraphics[width=0.2\columnwidth]{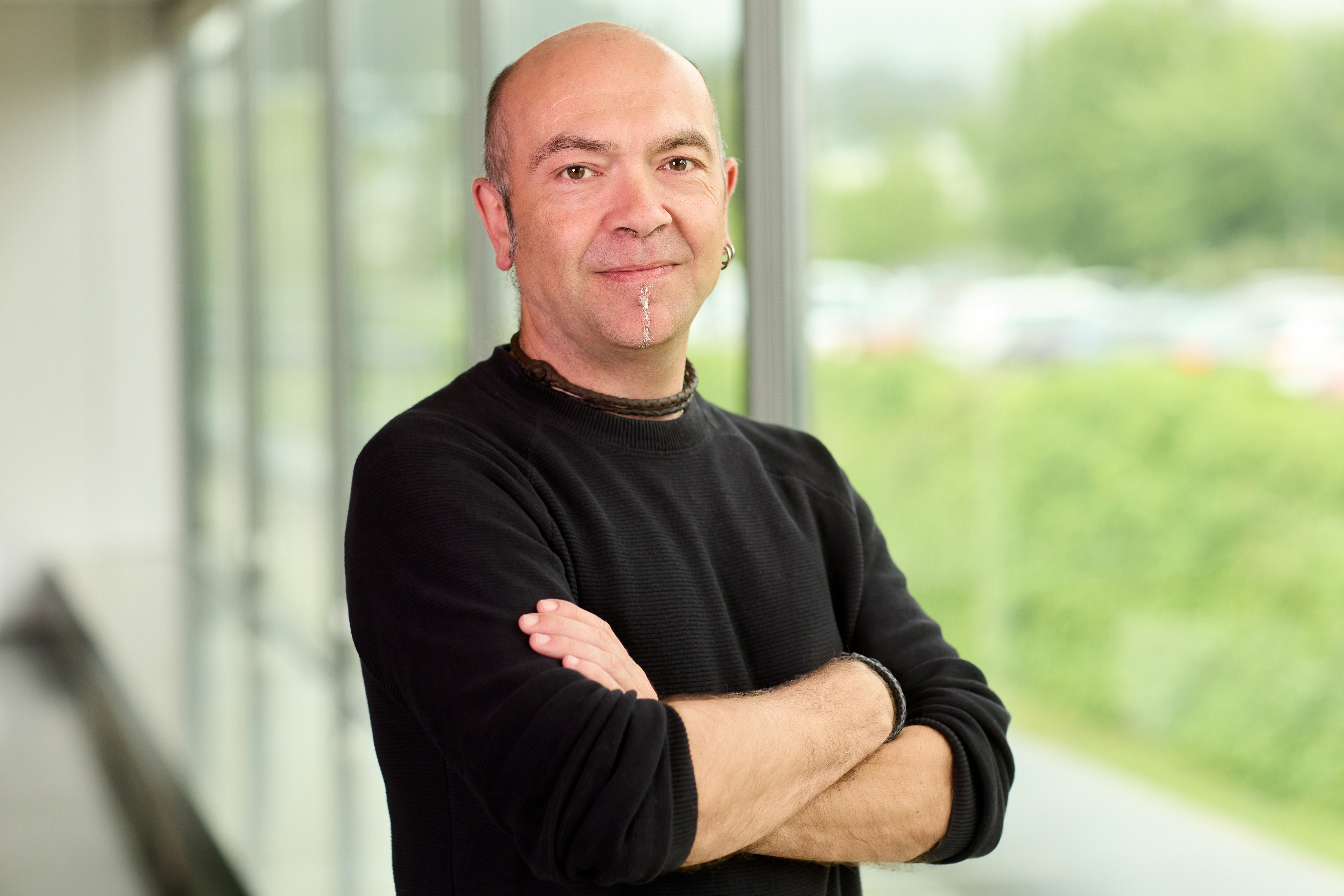}
\end{figure}
\textbf{Sergio Gil-Lopez} received the M.Sc. degree in Physics from Universidad de La Laguna, Tenerife, Spain in 2001, although the first half of his university studies was done in the Universidad Autónoma de Madrid, Spain. His Ph.D. degree in Atmospheric Physics from Universidad de Granada (at Instituto de Astrofísica de Andalucía, CSIC) was obtained in 2006. During his Ph. D. studies he spent three months in two Atmospheric German researcher centers (IMK in Karlsruhe and ICFG1 in Jüelich) for scientist collaboration. His thesis deals about Stratospheric to Mesospheric Ozone Retrieval from MIPAS/ENVISAT data. He was the head of the AI group of TECNALIA for 2 years as a Head of the most technological AI Group, and 5 years of the Head of AI strategy in TECNALIA. Nowadays, he continues being a Data Scientist senior researcher in TECNALIA RESEARCH \& INNOVATION with nearly 15 years of experience belonging to the Big Data TECNALIA’s Team. Although he has had one year’s break in Ariadna Instrument S. L. working in the developing of an Intelligent algorithm for Low Voltage Network Topology Estimation, fraud detection techniques and energy balance estimations for the Smart Grids. He has co-authored more than 40 international journal papers and more than 43 conference contributions, 7 filed patents and 2 supervised PhDs. He has also been involved in the organization of various national and international conferences, either in charge for chairing positions.

\vskip1pc

\begin{figure}[h!]
    \centering
    \includegraphics[width=0.12\columnwidth]{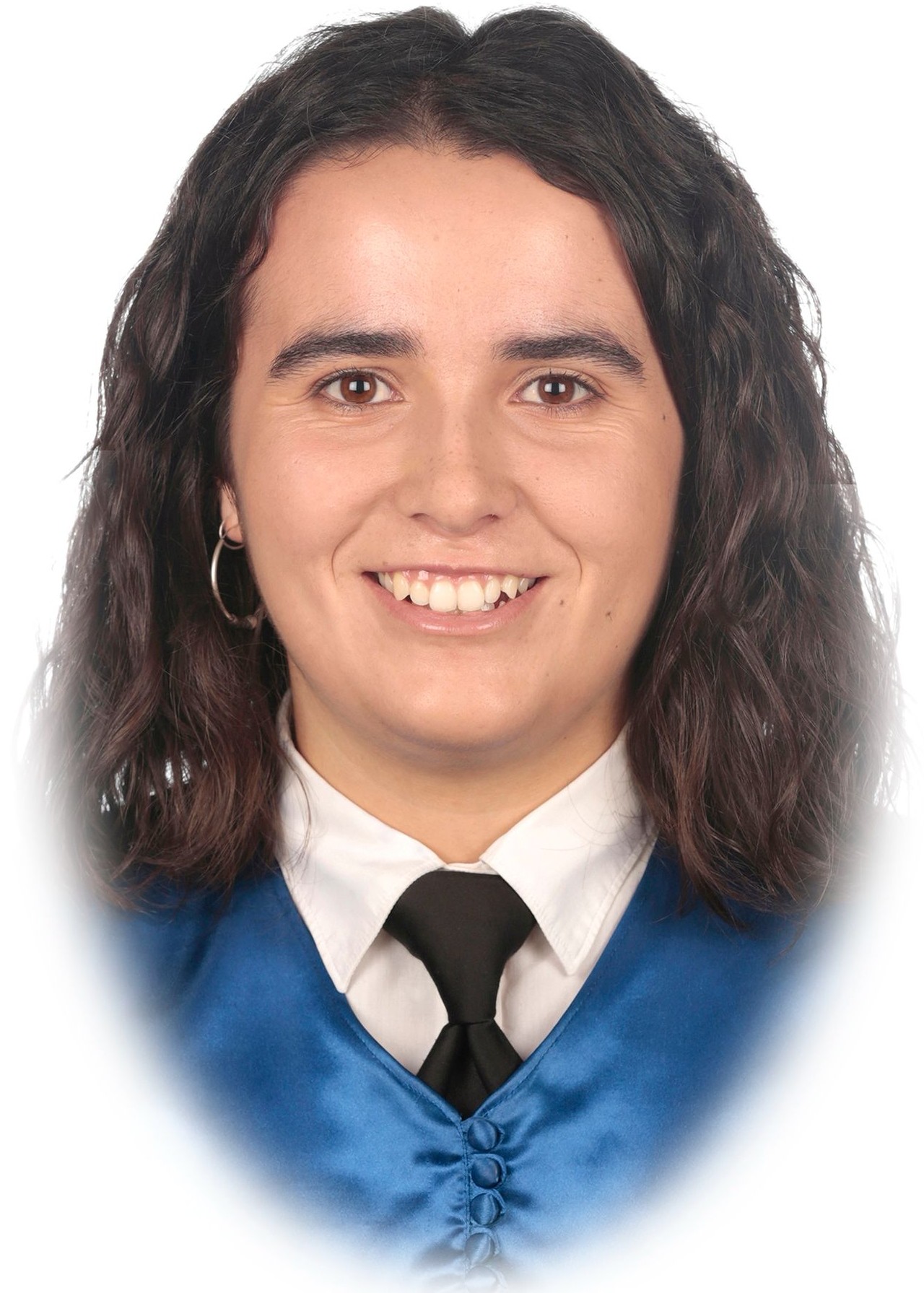}
\end{figure}
\textbf{Leire Arrizabalaga} obtained her B.Sc. degree in Mathematics from the University of the Basque Country in March 2026. During her final year, she focused on the present research in order to achieve results for her Bachelor's thesis. She has developed strong expertise in artificial intelligence and data science techniques, with a particular focus on the energy domain and, specifically, battery systems.

\vskip4pc

\begin{figure}[h!]
    \centering
    \includegraphics[width=0.2\columnwidth]{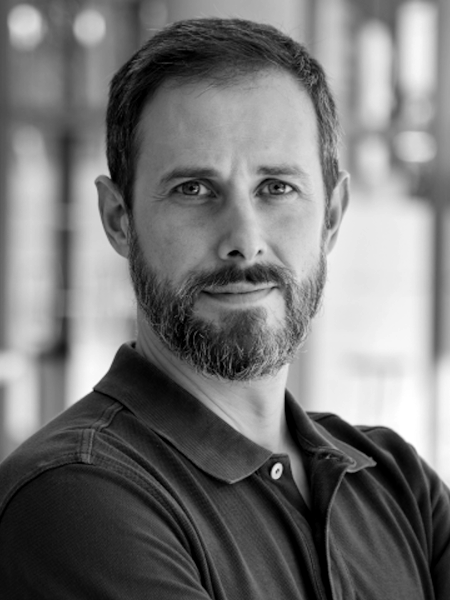}
\end{figure}
\textbf{Ibai Laña} received his B.Sc. degree in Computer Engineering from Deusto University, Spain, in 2006, the M.Sc. degree in Advanced Artificial Intelligence from UNED, Spain, in 2014, and the PhD in Artificial Inteligence from UPV/EHU in 2018. He is currently a researcher in the Mathmode group and a research professor at Mondragon University.  His research interests fall within the intersection of Intelligent Transportation Systems (ITS), machine learning, traffic data analysis and data science. He has dealt with urban traffic forecasting problems, where he has applied machine learning models and evolutionary algorithms to obtain longer term and more accurate predictions.  In recent years, his work has focused on modeling complex systems using graph neural networks (GNNs) and on developing explainability techniques (XAI) for artificial intelligence models. He has led technical AI projects and has extensive experience in heuristic optimization, predictive analytics, and operational efficiency.

\end{document}